\documentclass{article}

\PassOptionsToPackage{numbers, compress}{natbib}

\usepackage[main,final]{neurips_2026}

\usepackage[utf8]{inputenc} %
\usepackage[T1]{fontenc}    %
\usepackage{hyperref}       %
\usepackage{url}            %
\usepackage{booktabs}       %
\usepackage{amsfonts}       %
\usepackage{nicefrac}       %
\usepackage{microtype}      %
\usepackage{xcolor}         %
\usepackage{needspace}      %

\newif\ifrevisioncolor
\revisioncolorfalse
\definecolor{revisionblue}{RGB}{0,85,170}

\usepackage{amsthm}
\usepackage{aliascnt}
\newcommand{\sharedtheorem}[3]{%
  \newaliascnt{#1}{theorem}%
  \newtheorem{#1}[#1]{#2}%
  \aliascntresetthe{#1}%
  \expandafter\def\csname #1autorefname\endcsname{#3}%
}
\theoremstyle{plain}
\newtheorem{theorem}{Theorem}[section]

\sharedtheorem{proposition}{Proposition}{Proposition}
\sharedtheorem{lemma}{Lemma}{Lemma}
\sharedtheorem{corollary}{Corollary}{Corollary}
\theoremstyle{definition}
\sharedtheorem{definition}{Definition}{Definition}
\sharedtheorem{assumption}{Assumption}{Assumption}
\theoremstyle{remark}
\sharedtheorem{remark}{Remark}{Remark}

\usepackage{caption}
\usepackage{algorithm}
\usepackage{algpseudocode}
\usepackage{multirow}
\usepackage{amsmath}
\usepackage{amssymb}
\usepackage{svg}
\usepackage{tabularx}
\usepackage{xspace}
\usepackage{wrapfig}
\usepackage[most]{tcolorbox}
\usepackage[flushleft]{threeparttable}
\usepackage{xcolor}
\usepackage{comment}
\usepackage{float}

\newtcolorbox{takeawaybox}{
    colback=blue!3!gray!5,       %
    colframe=blue!40!black,      %
    boxrule=0.5pt,               %
    arc=1mm,                     %
    left=4mm,                    %
    right=2mm,
    top=2mm,
    bottom=2mm,
    enhanced,                    %
    borderline west={3pt}{0pt}{blue!60!cyan} %
}

\newcommand{\methodname}{\texttt{FEROM}\xspace}

\title{\methodname: Frontier Endogenous Reveal-Order Marginal Policy Optimization for Masked Diffusion LMs}
\hypersetup{
  pdftitle={FEROM: Frontier Endogenous Reveal-Order Marginal Policy Optimization for Masked Diffusion LMs},
  pdfauthor={Zian Su, Ziyang Huang, Kaiyuan Zhang, Xiangyu Zhang}
}

\author{
Zian Su \\
Purdue University \\
\texttt{su284@purdue.edu}
\And
Ziyang Huang \\
Johns Hopkins University \\
\texttt{zhuang86@jhu.edu}
\AND
Kaiyuan Zhang \\
Rutgers University \\
\texttt{kaiyuan.z@rutgers.edu}
\And
Xiangyu Zhang \\
Purdue University \\
\texttt{xyzhang@cs.purdue.edu}
}

\begin{document}

\maketitle

\begin{abstract}

Masked diffusion language models generate text by iteratively unmasking positions, with practical samplers often selecting reveal positions from the model's own logits. The reveal order is therefore an endogenous latent variable of the rollout process with an enormous discrete latent space. Policy optimization objectives relying on pre-defined schedules or sampled trajectories introduce either bias or high variance. In this paper, we propose \texttt{FEROM}, \emph{Frontier Endogenous Reveal-Order Marginal Policy Optimization}, which approximates policy gradients for the rollout-induced marginal response policy of masked diffusion LMs. \texttt{FEROM} derives a Rao--Blackwellized policy-gradient identity over latent reveal orders and expresses the resulting estimator as posterior edge occupancy on a reveal-state DAG. To make marginalization practical, we introduce Frontier Reveal Marginalization, a budgeted estimator that approximates this marginal edge structure by combining CDF-based computation of local probabilities of the endogenous reveal process with frontier expansion of high-mass partial states. Integrated into a GRPO-style objective, \texttt{FEROM} replaces single-path log-scores with a detached token-score surrogate over posterior-relevant reveal edges. Experiments on math and coding tasks show comparable or improved results over existing methods under matched scoring-state budgets. An offline proxy study further shows improved marginal-gradient approximation with increased scoring budgets.\footnote{Code is available at: \url{https://github.com/ziansu/ferom}.}
\end{abstract}

\section{Introduction}

Diffusion language models (dLLMs)~\cite{li2022diffusion,gong2022diffuseq} and, in particular, \emph{masked diffusion language models} (MDMs)~\cite{shi2024simplified,sahoo2025diffusion} have recently emerged as promising alternatives to left-to-right autoregressive (AR) language models. Instead of committing to a fixed causal factorization, an MDM generates text by iteratively refining a partially masked sequence. This decoding paradigm naturally supports bidirectional conditioning, adaptive reveal orders, and partially parallel token updates, leading to attractive latency---parallelism trade-offs at inference time~\cite{nie2026large,ye2025dream}.

This flexibility, however, creates a distinctive challenge for reinforcement learning. In an AR model, a sampled response comes with a unique token-generation trajectory, so the rollout policy and the policy log-probability used in policy-gradient updates are naturally aligned. In an MDM, generation proceeds through a sequence of partially revealed states, and the sampler must repeatedly decide which masked positions to reveal next. In practical systems, these reveal decisions are often computed from the model's own logits, for example, through confidence or margins~\cite{nie2026large,kim2025train}. Thus, the reveal order is not merely an external schedule: it is an \emph{endogenous} latent process coupled to token prediction.

On the other hand, sequence-level rewards evaluate only the completed response, not the internal reveal trace used to produce it. The same response may be reachable through many plausible reveal orders, all equivalent from the reward function's perspective. A trajectory-level policy-gradient estimator that conditions on the single sampled reveal path is valid for the joint rollout process, but it attributes reward to latent reveal-order randomness that the reward never observes.

In this paper, we argue that the statistically appropriate object for sequence-level RL is instead the {\emph{rollout-induced marginal response policy}}: the distribution over completed responses after integrating out reveal orders. Its policy gradient averages trajectory-level contributions over reveal traces consistent with the same completed response, a Rao--Blackwellization that removes nuisance pathwise variance. The difficulty is computational. Exact marginalization over endogenous reveal orders is combinatorial in sequence length, while MDM RL is already constrained by expensive transformer forwards, long denoising horizons, and limited GPU memory. This raises the central question of this paper: {\color{purple}\emph{how can we estimate the policy gradient of the rollout-induced marginal response policy under a realistic compute and memory budget?}}

We propose \methodname, \emph{Frontier Endogenous Reveal-Order Marginal Policy Optimization}, a policy-gradient framework for masked diffusion LMs under their actual rollout process. \methodname targets the rollout-induced marginal response policy: it first Rao--Blackwellizes trajectory policy gradients over latent reveal orders, then represents reveal trajectories as paths in a reveal-state DAG, where posterior edge occupancies weight each reveal transition's contribution to the marginal gradient. Since exact occupancy computation is combinatorial, \methodname introduces Frontier Reveal Marginalization (FRM). FRM approximates the marginal edge structure by maintaining a frontier of high-occupancy partial states, computing target-conditioned local probabilities of the endogenous reveal process exactly from confidence-score cumulative distribution functions (CDFs), and expanding posterior-relevant reveal edges under a fixed scoring-state budget. The resulting \methodname-GRPO objective uses this edge structure to replace the log-score of a single sampled reveal path with a detached token-score surrogate over posterior-relevant reveal edges.

We evaluate \methodname on mathematical reasoning and code generation using GSM8K~\cite{cobbe2021training}, MATH-500~\cite{lightman2024lets}, HumanEval~\cite{chen2021codex}, and MBPP~\cite{austin2021program}, with LLaDA-Instruct~\cite{nie2026large} as the base model.
At the 256-token training length, \methodname leads the compared small-budget baselines on MATH-500 and HumanEval and ranks second on GSM8K. Our offline study further shows that larger scoring budgets allow \methodname to better approximate the block-level marginal gradient than single-trajectory variants.

Our contributions are fourfold. First, we formulate MDM reinforcement learning as optimizing the rollout-induced marginal response policy, with reveal orders treated as endogenous latent variables generated by the model's logits. Second, we derive a Rao--Blackwellized marginal policy-gradient identity and characterize its estimator as posterior edge occupancy on a reveal-state DAG. Third, we introduce \methodname, a budgeted frontier-based algorithm that approximates the token-score component of this marginal gradient under strict forward and memory constraints via exact local probabilities of the endogenous reveal process and budgeted edge coverage. Fourth, we instantiate \methodname in a GRPO-style objective and demonstrate benefits in low-budget training and block-level marginal-gradient approximation.

\section{Preliminaries}
\label{sec:preliminaries}

\begin{figure}
    \centering
    \includegraphics[width=1.0\linewidth]{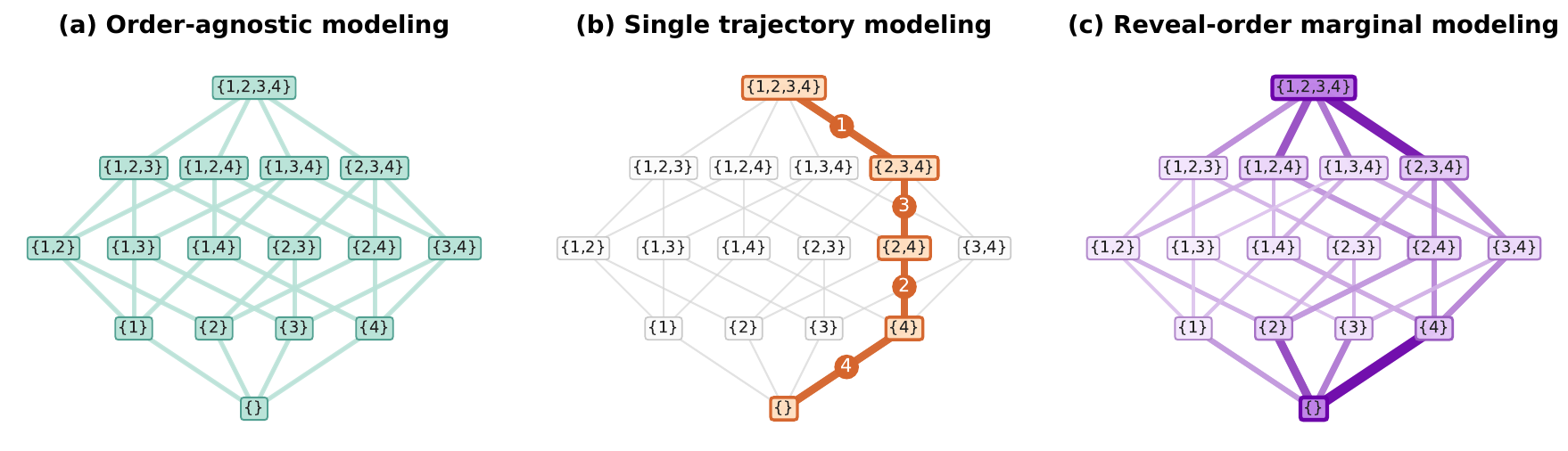}
  \caption{
\textbf{Reveal-order modeling on an example 4-position reveal-state DAG.}
Nodes show the still-masked sets \([n]\setminus S\), where \(S\) is the revealed set. Each edge reveals one position. Node color denotes state occupancy, and edge color/width denotes reveal-edge occupancy.
\textbf{(a)} Order-agnostic denoising uses training states from an exogenous masking process rather than the adaptive rollout distribution.
\textbf{(b)} Single-trajectory modeling follows one sampled reveal path, assigning credit only to visited edges and retaining latent reveal-order variance.
\textbf{(c)} \methodname aggregates reveal-order contributions using posterior edge occupancies on the DAG.%
}
    \label{fig:conceptual-poset}
\end{figure}

In this section, we first review masked diffusion model basics and then introduce an alternative endogenous reveal-process view that we advocate throughout this paper.

\paragraph{Notation.}
Let \(x^{1:n}\) denote a length-\(n\) token sequence, with \(x^i\in\mathcal{V}\) the token at position \(i\), and let \([n]\triangleq\{1,\dots,n\}\). \(\mathbf e_{x^i}\) is the one-hot vector of token \(x^i\), and \(\mathbf e_{\texttt{[M]}}\) is the one-hot vector of the mask token. For any subset \(S\subseteq[n]\), let \(\tilde x_S\) denote a partially revealed sequence, also referred to as a state, where \(S\) is the set of currently revealed positions and the remaining positions are all masks. We use \(\mathrm{Cat}(\cdot;p)\) for the categorical distribution with parameter \(p\in\Delta(\mathcal{V}\cup\{\texttt{[M]}\})\).

\paragraph{Forward process.}
For a clean sequence \(x_0\sim p_{\mathrm{data}}\) and noise level \(t\sim\text{Unif}[0,1]\), the forward corruption process \textit{independently} masks each token according to
$
q_{t|0}(x_t^{1:n}\mid x_0^{1:n})
=
\prod_{i=1}^n q_{t|0}(x_t^i\mid x_0^i),
$
with
\[
q_{t|0}(x_t^i\mid x_0^i)
=
\mathrm{Cat}\!\left(
x_t^i;\;
\alpha_t \mathbf e_{x_0^i}
+
(1-\alpha_t)\mathbf e_{\texttt{[M]}}
\right),
\]
where \(\alpha_t\) is a predefined noise schedule, commonly chosen as $\alpha_t=1-t$.

\paragraph{Reverse process and training.} The corresponding reverse process $q_{s|t}(x_s^{1:n}|x_t^{1:n},x_0^{1:n})=\prod_{i=1}^{n}q_{s|t}(x_s^i|x_t^i,x_0^i)$, where $s < t$, reverses the forward corruption. A time-agnostic neural network $p_\theta(x_0^i|x_t^{1:n})$ approximates this process by predicting the clean token $x_0^i$ given the randomly masked state $x_t^{1:n}$. Training minimizes the following negative evidence lower bound (NELBO)~\cite{zheng2024masked,shi2024simplified,sahoo2024simple}:

\begin{equation}
    \mathcal{L}_{\theta}\!=\!\int_0^1 \omega(t) \ \mathbb{E}_{\substack{x_0\sim p_{\mathrm{data}},\\x_t\sim q_{t|0}(\cdot|x_0)}}\left[\sum_{i: x_t^i=\texttt{[M]}}\!-\!\log p_\theta(x_0^i|x_t^{1:n})\right]\mathrm{d}t
\end{equation}

where $\omega(t)=\frac{-\alpha'_t}{1-\alpha_t}$ and $\alpha_t'=\frac{\mathrm{d} \alpha_t}{\mathrm{d} t} < 0$.
The standard MDM objective samples states from the prescribed forward corruption process. In RL, using this objective as a reward-weighted surrogate optimizes a masking-state distribution that may differ from the rollout distribution~\cite{wang2025revolutionizing,kim2026stop,turok2026duel}.

\paragraph{Endogenous reveal process.}
At inference, an MDM alternates between token prediction and reveal-position selection~\cite{sahoo2024simple,nie2026large,Bie2025LLaDA2}. A reveal rule is \emph{endogenous} when it selects positions using the same logits as token prediction. LLaDA-style decoding keeps revealed tokens fixed and resamples candidates from current-state predictions. Hence the current state determines the next-update distribution for a fixed context and decoder configuration.
For a one-token update,\footnote{Parallel multi-token updates use a joint subset-selection probability, derived in Appendix~\ref{app:local_order}.} let \(Z_j\) be the candidate at masked position \(j\) and \(I\) the selected position. We use \(p_\theta\) for the candidate-token distribution at the rollout temperature. For a sequence \(x\), define \(r_\theta^x(i\mid\tilde x_S)\triangleq\Pr_\theta(I=i\mid\tilde x_S,Z_i=x^i)\).
Only \(Z_i\) is fixed; competing candidates remain random. For a reveal order \(\sigma=(\sigma_1,\dots,\sigma_n)\), let \(S_0=\emptyset\) and \(S_t=\{\sigma_1,\dots,\sigma_t\}\). The chain rule gives the joint distribution
\begin{equation}
P_\theta(x,\sigma\mid\tilde x_{S_0})
=
\prod_{t=0}^{n-1}
r_\theta^x(\sigma_{t+1}\mid\tilde x_{S_t})
\;
p_\theta(x^{\sigma_{t+1}}\mid\tilde x_{S_t}).
\end{equation}

\begin{remark} 
Classical order-agnostic AR models marginalize over an exogenous order distribution, often uniform over permutations~\cite{uria2014deep,hoogeboom2021autoregressive,austin2021structured,zheng2024masked,ou2024your}. In contrast, an MDM's endogenous reveal process induces a state-dependent reveal distribution from the model's own predictions, so the latent order is part of the sampler whose marginal response policy is optimized.
\end{remark}

\section{\methodname: Frontier Endogenous Reveal-Order Marginal Policy Optimization}

We introduce \methodname through its marginal-gradient and reveal-state formulation (\S\ref{sec:theory}--\ref{sec:reveal_state_dag}), budgeted frontier approximation (\S\ref{sec:frm}), and resulting GRPO-style objective (\S\ref{sec:ferom_grpo}).

\subsection{Theoretical Foundations: Marginalizing Latent Reveal Traces}
\label{sec:theory}

We formalize reinforcement learning for MDMs under the endogenous reveal-process view. Let \(c\sim\rho\) be a conditioning context drawn from the prompt distribution \(\rho\), and \(x\) a generated sequence. Given a sequence-level reward \(R(c,x)\), the RL objective is
\[
J(\theta)=\mathbb E_{c\sim\rho,x\sim \pi_\theta(\cdot\mid c)}[R(c,x)].
\]

At first glance, a trajectory-level policy-gradient estimator that samples a full reveal trace \((x,\sigma)\sim P_\theta(\cdot,\cdot\mid c)\) and uses
\[
g_{\mathrm{traj}}(x,\sigma,c)=R(c,x)\nabla \log P_\theta(x,\sigma\mid c)
\]
is most straightforward since it is on-policy for the joint rollout process. However, this trajectory-level estimator conditions on the particular reveal trace sampled during rollout. Since for language models the reward $R(c,x)$ depends only on the completed sequence \(x\) and the context $c$, the reveal order is a nuisance latent variable for sequence-level credit assignment. Let
\(
\pi_\theta(x\mid c)=\sum_\sigma P_\theta(x,\sigma\mid c)
\)
denote the reveal-order-marginal policy induced by the rollout process. Differentiating the marginal policy yields
\[
\nabla \log \pi_\theta(x\mid c)
=
\mathbb E_{\sigma\sim P_\theta(\sigma\mid c,x)}
\!\left[
\nabla \log P_\theta(x,\sigma \mid c)
\right].
\]
The sequence-level score is the posterior average of trajectory scores for the same sequence, an instance of Fisher's identity. Substituting it into the policy-gradient objective gives
\begin{equation}
\begin{aligned}
\nabla J(\theta)
&=
\mathbb E_{\substack{c\sim\rho,\\x\sim \pi_\theta(\cdot\mid c)}}
\!\left[
R(c,x)\nabla \log \pi_\theta(x\mid c)
\right]
\\
&=
\mathbb E_{\substack{c\sim\rho,\\ x\sim \pi_\theta(\cdot\mid c)}}
\!\left[
R(c,x)\,
\mathbb E_{\sigma\sim P_\theta(\sigma\mid c,x)}
\!\left[
\nabla \log P_\theta(x,\sigma\mid c)
\right]
\right].
\end{aligned}
\end{equation}
This is the marginal policy gradient targeted by \methodname.

\begin{theorem}[Rao-Blackwell dominance of marginal policy gradients]
\label{thm:rb_policy_gradient}
Assume that the reward depends only on the completed sequence \(x\) and context \(c\). Assume also that differentiation can be interchanged with the trajectory sum and that the scores have finite second moments. Define
\[
g_{\mathrm{traj}}(x,\sigma,c)
=
R(c,x)\nabla\log P_\theta(x,\sigma\mid c),
\qquad
g_{\mathrm{marg}}(x,c)
=
R(c,x)\nabla\log \pi_\theta(x\mid c).
\]
Then,
    (1) \(\mathbb E[g_{\mathrm{traj}}]=\mathbb E[g_{\mathrm{marg}}]=\nabla J(\theta)\);
    (2) \(g_{\mathrm{marg}}(x,c)=\mathbb E[g_{\mathrm{traj}}(x,\sigma,c)\mid x,c]\);
    (3) \(\operatorname{Cov}(g_{\mathrm{marg}})\preceq \operatorname{Cov}(g_{\mathrm{traj}})\);
    (4) equality in (3) holds iff \(\operatorname{Var}(g_{\mathrm{traj}}\mid x,c)=0\) almost surely.
\end{theorem}

Theorem~\ref{thm:rb_policy_gradient} motivates averaging over reveal orders: exact posterior averaging preserves the expected policy gradient and reduces reveal-order variance (proof in Appendix~\ref{appx:theory}).
 
\begin{takeawaybox}
    \textbf{Takeaway.}
    The reward observes the completed sequence, not its internal reveal trace. Averaging over traces consistent with that sequence removes this source of gradient variance.
\end{takeawaybox}

\begin{figure}[!t]
    \centering
    \includegraphics[width=\linewidth]{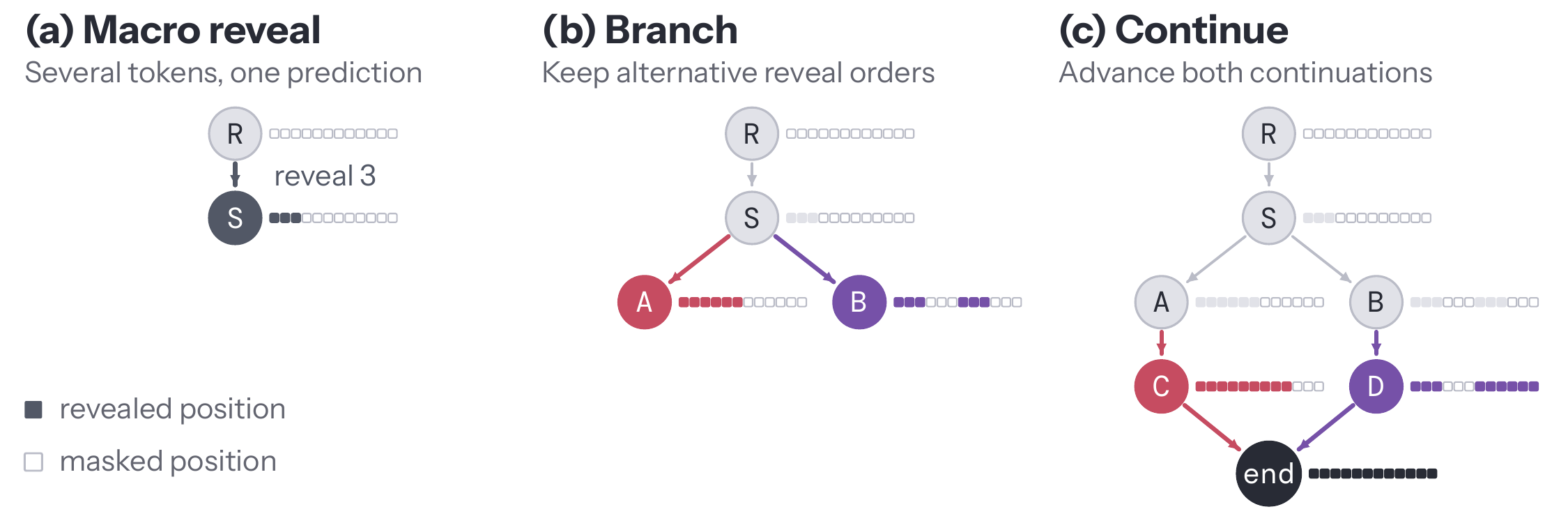}
    \caption{\textbf{Macro reveals and branching in FRM.} FRM progressively expands the reveal-state DAG for a given sequence: \textbf{(a)} a macro reveal scores several target tokens from one parent prediction; \textbf{(b)} branching creates two continuations with different leading reveals; \textbf{(c)} further macro reveals advance both continuations to the same completed sequence. Filled cells mark revealed positions; colored arrows show newly added transitions.}
    \label{fig:ferom-planning}
\end{figure}

\subsection{Reveal-State DAG Representation and Posterior Edge Occupancy}
\label{sec:reveal_state_dag}

Theorem~\ref{thm:rb_policy_gradient} motivates averaging over reveal trajectories, but computing this average directly requires considering combinatorially many reveal orders. To develop a budgeted approximation, we first rewrite the marginal score on a \emph{reveal-state directed acyclic graph (DAG)}. This representation lets us collect contributions from different orders through the transitions they share.

\paragraph{Reveal-state DAG.} We define a DAG given a context \(c\) and a completed sequence \(x\). Here, each node is a revealed set \(S\subseteq[n]\), and each edge
\(
    e=(S,i), i\notin S,
\)
reveals position \(i\) and moves from \(S\) to \(S\cup\{i\}\). A complete path from \(\emptyset\) to \([n]\) corresponds to one reveal order. For an order \(\sigma=(\sigma_1,\ldots,\sigma_n)\), write
\(
    S_t(\sigma)=\{\sigma_1,\ldots,\sigma_t\}
\)
for the revealed set after \(t\) steps, with \(S_0(\sigma)=\emptyset\). The next edge has \(S=S_t(\sigma)\) and \(i=\sigma_{t+1}\), so the trajectory's traversed edge set is
\(
    \mathcal E(\sigma)
    =
    \{(S_t(\sigma),\sigma_{t+1}):t=0,\ldots,n-1\}.
\)
Orders that reach the same revealed set share the same partially revealed sequence and its possible continuations. We illustrate how different reveal-order modeling approaches use this structure in Figure~\ref{fig:conceptual-poset} and provide a concrete example in Appendix~\ref{appx:example-oracle-marginal}.

\paragraph{Posterior edge occupancy.} The reveal orders consistent with \(x\) are weighted by their posterior \(P_\theta(\sigma\mid c,x)\). Their combined probability of traversing an edge \(e\) defines its \emph{posterior edge occupancy}:
\[
    \mu_{\theta,e}^\star(c,x)
    =
    \Pr_{\sigma\sim P_\theta(\sigma\mid c,x)}
    \left[
        e\in\mathcal E(\sigma)
    \right].
\]
Since each trajectory score is a sum of local edge scores, averaging over trajectories weights each edge by its posterior occupancy. For differentiable local edge factors, this gives the \emph{exact marginal score}:

\begin{equation}
    \label{eq:exact-marginal-gradient}
    \nabla_\theta \log \pi_\theta(x\mid c)
    =
    \sum_{e=(S,i)}
    \mu_{\theta,e}^\star(c,x)
    \nabla_\theta
    \log
    \Bigl[
        r_\theta^x(i\mid c,\tilde x_S)
        p_\theta(x^i\mid c,\tilde x_S)
    \Bigr].
\end{equation}

Equation~\ref{eq:exact-marginal-gradient} averages both token-prediction and reveal-decision scores over posterior edge occupancies. Appendix~\ref{app:exact_dag_dp} derives these occupancies and the identity.

\paragraph{Local reveal probabilities for LLaDA decoding.}
The selection factor \(r_\theta^x\) in Eq.~\ref{eq:exact-marginal-gradient} depends on how the decoder chooses among sampled candidates. For a one-token LLaDA-style update, let \(\mathcal I(S)\) be the eligible masked positions: \([n]\setminus S\), restricted to the active block for blockwise decoding. At position \(j\), logits \(\mathbf a_{\theta,j}=\operatorname{logits}_\theta(c,\tilde x_S)_j\) give the candidate distribution \(p_{\theta,j}=\operatorname{softmax}(\mathbf a_{\theta,j}/\tau)\) at temperature \(\tau>0\). The decoder independently samples \(Z_j\sim p_{\theta,j}\) at each eligible position and reveals the candidate with the highest untempered confidence \(\operatorname{softmax}(\mathbf a_{\theta,j})_{Z_j}\).

For \(i\in\mathcal I(S)\), condition on \(Z_i=x^i\). The selection probability \(r_\theta^x(i\mid c,\tilde x_S)\) is the probability that no competing candidate outranks this fixed target. The confidence cumulative distribution functions (CDFs) \(F_{\theta,j}^{\le}(u)\) and \(F_{\theta,j}^{<}(u)\) give the probabilities of confidence at most \(u\) and strictly below \(u\), respectively. Independence gives the following product, with lower position indices winning ties:
\begin{equation}
    \label{eq:local-cdf-selection}
    r_\theta^x(i\mid c,\tilde x_S)
    =
    \prod_{\substack{j\in\mathcal I(S)\\j<i}}
        F_{\theta,j}^{<}\!\left(\operatorname{softmax}(\mathbf a_{\theta,i})_{x^i}\right)
    \prod_{\substack{j\in\mathcal I(S)\\j>i}}
        F_{\theta,j}^{\le}\!\left(\operatorname{softmax}(\mathbf a_{\theta,i})_{x^i}\right).
\end{equation}
Each CDF is a vocabulary sum of known candidate probabilities, requiring no sampling or additional model forward. The values \(r_\theta^x\) need not sum to one over positions because each conditions on a different target event. Appendix~\ref{app:local_order} gives the \(k\)-token generalization and computational details.

\begin{algorithm}[!t]
\captionsetup{font=small}
\caption{Frontier Reveal Marginalization (FRM)}
\label{alg:frm}
\begin{algorithmic}[1]
\small
\Require context \(c\), sequence \(x\), initial state \(S_0\), model \(p_\theta\), state-evaluation budget \(B\)
\State \(\mathcal F\leftarrow\{S_0\}\), \(\widehat\nu(S_0)\leftarrow1\), \(\mathcal E_B\leftarrow\emptyset\), \(b\leftarrow0\)
\While{\(b<B\) and \(\mathcal F\neq\emptyset\)}
    \State Select and remove a continuation at \(S\) from \(\mathcal F\)
    \State Evaluate the model at \((c,\tilde x_S)\); \(b\leftarrow b+1\)
    \State Compute \(p_{\mathrm{loc},\theta}\) by Eq.~\ref{eq:frm-local-target}; condition on allowed leading positions
    \State Choose macro size \(k\) using remaining positions and the pre-split budget, including this evaluation
    \If{alternatives warrant branching and the budget supports both branches}
        \State \(i\leftarrow\arg\max_j p_{\mathrm{loc},\theta}(j\mid S,c,x)\)
        \State Reserve the complementary continuation at \(S\), excluding \(i\); divide the future budget
    \Else
        \State \(i\sim p_{\mathrm{loc},\theta}(\cdot\mid S,c,x)\); retain this continuation's remaining budget
    \EndIf
    \State Draw \(\Delta S\), with \(|\Delta S|=k\), conditional on leading reveal \(i\)
    \State \(e\leftarrow(S,i,\Delta S)\); assign \(\widehat\mu_e\) by Eq.~\ref{eq:frm-contribution-rule}
    \State Compute \(\ell_{\theta,\Delta S}(S;c,x)\); append \(e\) with its score and coefficient to \(\mathcal E_B\)
    \State Propagate \(\widehat\mu_e\) to \(S\cup\Delta S\) by Eq.~\ref{eq:frm-child-contribution}
    \State Update \(\mathcal F\) with the unfinished child and reserved alternatives
\EndWhile
\State \Return \(\mathcal E_B\) and \(\{\widehat\mu_e\}_{e\in\mathcal E_B}\)
\end{algorithmic}
\end{algorithm}

\Needspace{9\baselineskip}
\subsection{Frontier Reveal Marginalization for Budgeted Estimation}
\label{sec:frm}

Directly applying Eq.~\ref{eq:exact-marginal-gradient} requires evaluating exponentially many reveal states, with costly model forward and backward passes at each state. We introduce \emph{Frontier Reveal Marginalization} (FRM) to couple planning with posterior occupancy estimation under limited computation, where a \emph{frontier} of partially revealed states expands through \emph{macro reveals} that score several tokens from one prediction and \emph{branching} that explores alternative reveal orders with appreciable probability (Figure~\ref{fig:ferom-planning}).

\paragraph{Local posterior approximation.}
At a revealed state \(S\), the \emph{parent state} of its outgoing transitions, FRM uses the local edge factors from \S\ref{sec:reveal_state_dag} to prioritize reveal choices likely to have produced the given sequence \(x\). Each choice must both propose the required token and select its position. Equation~\ref{eq:local-cdf-selection} gives the selection probability conditional on that proposal. Multiplying by the probability of proposing \(x^i\) and normalizing across eligible positions gives
\begin{equation}
    \label{eq:frm-local-target}
    p_{\mathrm{loc},\theta}(i\mid S,c,x)
    =
    \frac{
        r_\theta^x(i\mid c,\tilde x_S)
        p_\theta(x^i\mid c,\tilde x_S)
    }{
        \sum_{j\in\mathcal I(S)}
        r_\theta^x(j\mid c,\tilde x_S)
        p_\theta(x^j\mid c,\tilde x_S)
    }.
\end{equation}
This distribution conditions on the next revealed token matching \(x\). FRM uses this one-step distribution as a local approximation to the next-reveal posterior conditioned on the full sequence, guiding position selection and branching from current-state predictions (Appendix~\ref{app:exact_dag_dp}).

\paragraph{Macro reveals: better coverage with fewer states.}
We measure the budget \(B\) in \emph{state evaluations}. A \emph{macro reveal} \(\Delta S\subseteq\mathcal I(S)\) uses one evaluation at parent state \(S\) to score the target tokens in this set, with log-score \(\ell_{\theta,\Delta S}(S;c,x)=\sum_{i\in\Delta S}\log p_\theta(x^i\mid c,\tilde x_S)\). To form the set, FRM selects a \emph{leading position} \(i\) using \(p_{\mathrm{loc},\theta}\) and fixes its candidate to \(x^i\). It samples competing candidates conditional on \(i\) ranking first by confidence; the highest-ranked positions, including \(i\), form \(\Delta S\). FRM records the macro-edge as \(e=(S,i,\Delta S)\) and advances to \(S\cup\Delta S\). Macro size adapts to the remaining masked positions and allocated state evaluations: fewer evaluations require larger reveals, while more evaluations permit finer state resolution. Appendix~\ref{app:frm_variants} specifies the proposal and size-selection rules.

\paragraph{Branching and frontier allocation.}
The frontier \(\mathcal F\) stores unfinished continuations of reveal paths for the same sequence \(x\). Each continuation specifies a current state \(S\), the allowed leading positions for its next macro reveal, and a remaining evaluation budget. It also carries a nonnegative weight \(\widehat\nu(S)\), representing its share of reveal-order probability under FRM's local approximation. Initially, the frontier contains only \(S_0\), with weight one and all eligible leading positions allowed.

At a state \(S\), FRM compares the most likely leading position \(i\) with the other allowed choices. If these alternatives carry appreciable probability and the remaining budget can advance both branches, FRM splits the continuation \emph{before sampling a macro-set}. One branch fixes \(i\) as its leading reveal; the other remains at \(S\) with \(i\) excluded from its allowed leading positions. The two branches divide the remaining state evaluations and advance through their own macro reveals. Otherwise, FRM samples one leading position and follows a single continuation. Expansion priority favors larger continuation weights and less concentrated local reveal distributions.

\paragraph{Occupancy estimation and contribution weights.}
Following the weighted-sum structure of Eq.~\ref{eq:exact-marginal-gradient}, FRM assigns each scored macro-edge a contribution coefficient \(\widehat\mu_e\). Here \(p_{\mathrm{loc},\theta}\) is Eq.~\ref{eq:frm-local-target}, restricted and renormalized over the continuation's allowed leading positions. Without a split, sampling from this distribution represents the alternatives through their sampling frequencies, so the sampled macro-edge inherits the full continuation weight. A split instead represents the selected leading position and its complement separately, so the selected branch receives only its probability share:
\begin{equation}
    \label{eq:frm-contribution-rule}
    \widehat\mu_e=
    \begin{cases}
        \widehat\nu(S),
        & i\text{ sampled without a split},\\
        \widehat\nu(S)\,p_{\mathrm{loc},\theta}(i\mid S,c,x),
        & i\text{ fixed by a split}.
    \end{cases}
\end{equation}
On a split, the complementary continuation at \(S\) retains weight \(\widehat\nu(S)-\widehat\mu_e\). Each scored macro-edge passes its coefficient to the child state:
\begin{equation}
    \label{eq:frm-child-contribution}
    \widehat\nu(S\cup\Delta S)
    \leftarrow\widehat\nu(S\cup\Delta S)+\widehat\mu_e.
\end{equation}
FRM returns the macro-edge records \(\mathcal E_B(c,x)\) and their coefficients for the token-score objective in \S\ref{sec:ferom_grpo}. Algorithm~\ref{alg:frm} summarizes the procedure; Appendix~\ref{app:frm_variants} details the sampling and allocation rules.

\begin{takeawaybox}
    \textbf{Takeaway.}
    Under a limited state-evaluation budget, FRM balances \emph{token coverage along each continuation} through macro reveals with \emph{coverage of alternative reveal orders} through selective branching.
\end{takeawaybox}

\subsection{\methodname-GRPO}
\label{sec:ferom_grpo}
We use the FRM records to construct a token-score surrogate for GRPO~\cite{guo2025deepseek}. For each context \(c\sim\rho\), the fixed rollout policy samples \(G\) responses,
\(
    x_g\overset{\mathrm{i.i.d.}}{\sim}
    \pi_{\theta_{\mathrm{old}}}(\cdot\mid c),\quad g=1,\ldots,G,
\)
and their sequence-level rewards determine group-relative advantages \(\{\widehat A_g\}_{g=1}^G\). FRM supplies a collection of macro-edge records \(\mathcal E_B(c,x_g)\) and contribution coefficients \(\widehat\mu_e\) for each response. These define the surrogate log-score
\begin{equation}
    \widehat L_\theta(c,x_g)
    =
    \sum_{e=(S,i,\Delta S)\in\mathcal E_B(c,x_g)}
    \operatorname{sg}(\widehat\mu_e)
    \ell_{\theta,\Delta S}(S;c,x_g),
\end{equation}
where \(\operatorname{sg}(\cdot)\) denotes stop-gradient. The selected edges and coefficients remain fixed within an update, while gradients flow through the token scores. This uses the token-score component of Eq.~\ref{eq:exact-marginal-gradient}; Appendix~\ref{app:exact_dag_dp} gives its error decomposition relative to the full marginal score.

\paragraph{Policy objective.}
Using the same FRM records and coefficients for both policies, define the surrogate score ratio
\(
    \widehat\rho_\theta(c,x_g)
    =\exp\!\left(
        \widehat L_\theta(c,x_g)
        -\widehat L_{\theta_{\mathrm{old}}}(c,x_g)
    \right).
\)
The resulting clipped objective is
\begin{equation}
    \mathcal J_{\methodname}(\theta)
    =
    \mathbb E
    \left[
    \frac{1}{G}
    \sum_{g=1}^G
    \min\!\left(
        \widehat\rho_\theta(c,x_g)\widehat A_g,\;
        \operatorname{clip}
        \bigl(\widehat\rho_\theta(c,x_g),1-\epsilon,1+\epsilon\bigr)
        \widehat A_g
    \right)
    \right]
    -\lambda_{\mathrm{KL}}\widehat D_{\mathrm{ref}}.
\end{equation}
The expectation is over \(c\sim\rho\) and the sampled response groups; \(\epsilon>0\) is the clipping threshold, \(\lambda_{\mathrm{KL}}\ge0\) is the regularization coefficient, and \(\widehat D_{\mathrm{ref}}\) is the reference-policy regularizer evaluated on the same FRM records.

\section{Experiments}

\paragraph{Setup.}
\label{sec:experimental_setup}
We evaluate \methodname on mathematical reasoning and code generation using LLaDA-Instruct~\cite{nie2026large} as the base model. Following the standard LLaDA-Instruct RL evaluation protocol~\cite{ou2025principled,ni2026flexibility}, we report results on GSM8K~\cite{cobbe2021training}, MATH-500~\cite{lightman2024lets}, HumanEval~\cite{chen2021codex}, and MBPP~\cite{austin2021program}. For training, we use GSM8K and the math training set for reasoning, and the AceCoder-87K~\cite{zeng-etal-2025-acecoder} hard split for code generation. We follow the small-budget setting of ESPO~\cite{ou2025principled}. Our main-table runs use 4-bit QLoRA with bfloat16 mixed precision. Since LLaDA uses semi-autoregressive blockwise decoding, we preserve the native block order and apply \methodname independently within each block. We use the macro-reveal framework described in \S\ref{sec:frm}; Appendix~\ref{app:experimental-protocols} provides training and evaluation details. For evaluation, generation lengths of 128, 256, and 512 tokens use 128, 256, and 512 decoding steps, respectively, matching the baselines.

\begin{table*}[t]
    \centering
    \caption{\textbf{Accuracy (\%) across sequence lengths.} Best is in bold; second-best is underlined.}
    \label{tab:main_results}
    \setlength{\tabcolsep}{4.5pt}
    \renewcommand{\arraystretch}{1.15}
    \small
    \begin{tabular}{l*{12}{c}}
    \toprule
    \multirow{2}{*}{\textbf{Model / Seq Len}}
    & \multicolumn{3}{c}{\textbf{GSM8K}}
    & \multicolumn{3}{c}{\textbf{MATH-500}}
    & \multicolumn{3}{c}{\textbf{HumanEval}}
    & \multicolumn{3}{c}{\textbf{MBPP}} \\
    \cmidrule(lr){2-4}\cmidrule(lr){5-7}\cmidrule(lr){8-10}\cmidrule(lr){11-13}
    & \textbf{128} & \textbf{256} & \textbf{512}
    & \textbf{128} & \textbf{256} & \textbf{512}
    & \textbf{128} & \textbf{256} & \textbf{512}
    & \textbf{128} & \textbf{256} & \textbf{512} \\

    \midrule
    \color{gray}LLaDA-8B-Instruct~\cite{nie2026large}
                    & \color{gray}68.7 & \color{gray}76.7 & \color{gray}78.2 %
                    & \color{gray}26.0 & \color{gray}32.4 & \color{gray}36.2 %
                    & \color{gray}27.4 & \color{gray}35.5 & \color{gray}37.8 %
                    & \color{gray}36.2 & \color{gray}41.2 & \color{gray}40.4 %
                    \\
    \color{gray}LLaDA-1.5~\cite{zhu2025llada}
                    & \color{gray}- & \color{gray}83.3 & \color{gray}- %
                    & \color{gray}- & \color{gray}- & \color{gray}- %
                    & \color{gray}\underline{29.3} & \color{gray}39.6 & \color{gray}\textbf{51.9} %
                    & \color{gray}39.6 & \color{gray}39.9 & \color{gray}38.8 %
                    \\
    \midrule
    d1~\cite{zhao2026d}
                    & 73.2 & 81.1 & 82.1 %
                    & 33.8 & 38.6 & 40.2 %
                    & - & - & - %
                    & - & - & - %
                    \\
    wd1~\cite{tang2025wd1}
                    & 77.2 & 80.8 & 82.3 %
                    & 33.3 & 34.4 & 39.0 %
                    & - & - & - %
                    & - & - & - %
                    \\
    $d$-TreeRPO~\cite{pan2025d}
                    & - & 81.2 & 82.6 %
                    & - & 37.7 & 38.1 %
                    & - & - & - %
                    & - & - & - %
                    \\
    ESPO~\cite{ou2025principled}
                    & \underline{80.0} & 82.3 & 83.7 %
                    & \underline{36.0} & 39.0 & \underline{43.4} %
                    & 28.1 & \underline{42.1} & \underline{50.0} %
                    & \textbf{47.4} & \underline{44.6} & \underline{44.2} %
                    \\
    GDPO~\cite{rojas2025improving}
                    & 78.4 & 82.8 & \underline{85.0} %
                    & 33.2 & 39.6 & 41.4 %
                    & 26.2 & 39.6 & 39.0 %
                    & \underline{43.6} & \textbf{50.6} & \textbf{47.1} %
                    \\
    SPG~\cite{wang2025spg}
                    & 78.5 & \textbf{86.1} & 84.5 %
                    & 33.4 & \underline{40.0} & 41.8 %
                    & - & - & - %
                    & - & - & - %
                    \\
    \midrule
    \methodname (Ours)
                    & \textbf{81.0} & \underline{85.7} & \textbf{85.5} %
                    & \textbf{38.5} & \textbf{43.9} & \textbf{46.0} %
                    & \textbf{36.6} & \textbf{46.3} & 48.2 %
                    & 38.2 & 44.0 & 39.2 %
                    \\

    \bottomrule
    \end{tabular}
    \end{table*}

\paragraph{Main results.}
\label{sec:main_results}

Table~\ref{tab:main_results} reports results for \methodname under the ESPO-matched small-budget setting. At 256 tokens, the training length, \methodname leads the compared methods on MATH-500 and HumanEval and ranks second on GSM8K. At 128 tokens, it leads on GSM8K, MATH-500, and HumanEval; at 512, it leads on both math benchmarks but scores below ESPO on HumanEval. \methodname scores below ESPO and GDPO on MBPP. We hypothesize that LLaDA-style confidence-based decoding limits the diversity of high-reward rollouts on this task. We leave co-designing decoding and endogenous reveal-order marginal policy optimization to future work. We additionally compare full-parameter \methodname with JustGRPO, achieving similar math and HumanEval results at half the scoring budget (Appendix~\ref{app:justgrpo-comparison}).

\Needspace{12\baselineskip}
\section{Analysis}

\subsection{Offline Approximation Quality}
\label{sec:proxy-offline}
\begin{figure}[tb]
    \centering
    \includegraphics[width=\linewidth]{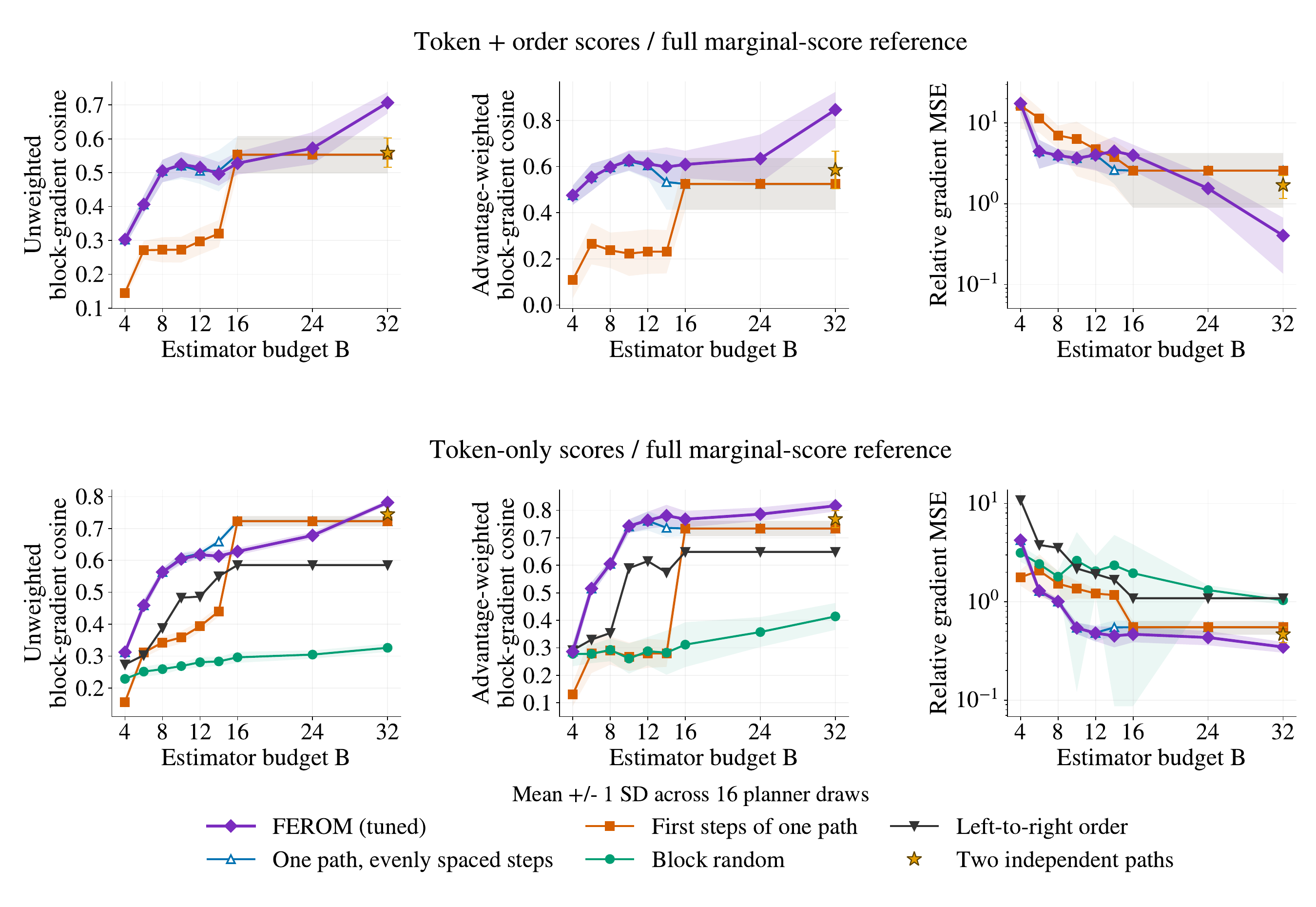}
    \caption{\textbf{Budgeted marginal-gradient approximation on AceCoder.} Token-plus-order (top) and token-only (bottom) estimates share a conditional marginal-gradient reference over 16 remaining positions. FEROM is tuned per budget on the displayed cohort, using one configuration across rows and metrics. Lines and bands show means and one standard deviation over 16 planner draws.}
    \label{fig:proxy-eval-approx}
\end{figure}

End-to-end RL combines approximation error with rollout and reward noise. We isolate estimation quality at continuation states from AceCoder rollouts, marginalizing the remaining reveal orders with conditional-CDF transition probabilities. Figure~\ref{fig:proxy-eval-approx} compares budgeted estimates against this full conditional marginal-score reference, with and without reveal-order derivatives on the same scored edges. Appendix~\ref{app:offline-protocol} gives the matched cases, gradient projection, and budget accounting.

\paragraph{Candidates.}
We compare FRM with one sampled path, a dense prefix, left-to-right order, block-random masking, and two independent paths when affordable. The single-path baseline evenly spaces evaluations along a macro-reveal chain; the dense prefix scores only its first one-token transitions. FRM allocates evaluations across alternative continuations. No candidate uses oracle occupancies.

\paragraph{Results.}
At small budgets, macro reveals provide coverage of the remaining sequence that a short dense prefix misses; FRM retains this benefit along a single path. In the token-only comparison, block-random masking has weaker gradient alignment across budgets, suggesting that broad mask coverage alone does not adequately represent the endogenous reveal process. Left-to-right ordering improves on random masking but generally trails model-induced path sampling, highlighting the importance of matching the reveal-order distribution. As the budget grows, frontier expansion improves advantage-weighted agreement beyond a single path; at the largest budget, it also improves agreement and lowers error relative to two independent paths. This pattern is consistent with probability concentrated around a dominant reveal path: a single path captures much of the marginal contribution, while branching uses additional budget to cover alternative orders.

\Needspace{5\baselineskip}
\subsection{Low-Budget Training Dynamics}

\begin{figure}[tb]
    \centering
    \includegraphics[width=\linewidth]{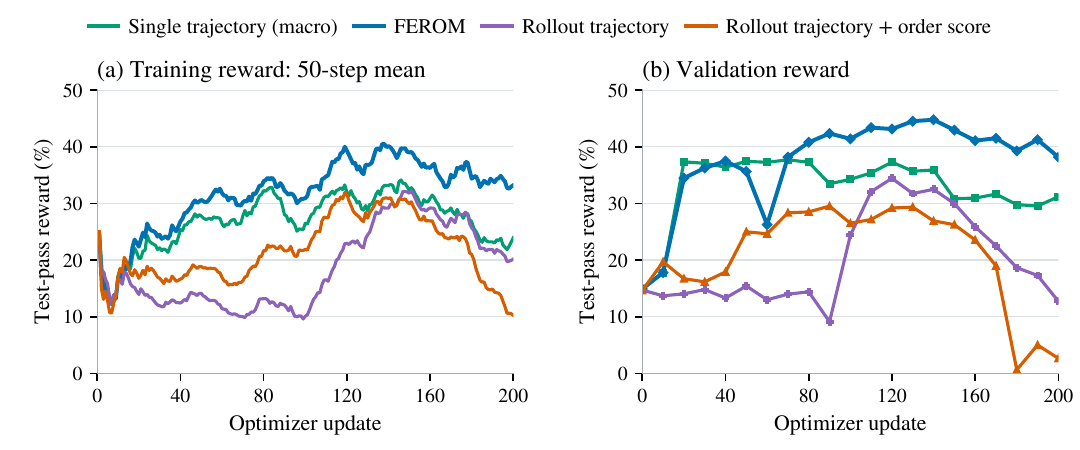}
    \caption{\textbf{Low-budget AceCoder training.} Training reward (left) and validation reward (right) for the frontier branching and reveal-order scoring ablations.}
    \label{fig:up2-stability}
\end{figure}

We use only two unique prompts per update, with eight responses each, to create a high-variance setting under a half-trajectory scoring budget. These full-parameter AceCoder experiments separately ablate branching and reveal-order scoring (Figure~\ref{fig:up2-stability}; Appendix~\ref{app:branching-protocol}).

Unlike \emph{Rollout trajectory}, \emph{Single trajectory (macro)} constructs a fresh reveal path for the generated sequence using FRM's local probabilities. Macro reveals also change the scoring resolution, while \methodname additionally branches across alternative continuations.

In the branching ablation, \methodname improves training and validation rewards over \emph{Single trajectory (macro)}, which uses the same macro-reveal horizon but disables branching.

In the reveal-order scoring ablation, \emph{Rollout trajectory} uses token scores at sampled rollout states; \emph{Rollout trajectory + order score} also differentiates the conditional reveal-order probability. The latter collapses early in training and validation, while the token-score control does not. This observed instability motivates the token-only surrogate for finite-budget training.

\Needspace{9\baselineskip}
\subsection{Computational Overhead}
Figure~\ref{fig:time-usage} compares CDF-based FEROM with a matched block-random control, separating model computation from CDF evaluation, macro-proposal sampling, and frontier bookkeeping. Model computation dominates the measured update, while CDF evaluation contributes a small fraction. In the memory-efficient replay implementation, separate planning and scoring forwards account for most of the additional cost. Sharing these forwards reduces repeated model computation when activation memory permits. We discuss more profiling details in Appendix~\ref{app:timing-protocol}.

\vspace{-0.3cm}
\section{Related Work}
\label{sec:related_work}

\paragraph{Train-test mismatch of unmasking process in MDMs.}
Concurrent work observes that MDM generation depends on the unmasking process, so random-mask training or ELBO evaluation can mismatch the states visited by practical samplers. DUEL~\citep{turok2026duel} formalizes deterministic position-selection rules and computes exact likelihood along the induced unmasking path, while PUMA~\citep{kim2026stop} reduces supervised train--test mismatch through progressive, teacher-forced unmasking. Our work differs in objective and estimator: \methodname targets reinforcement learning. Rather than evaluating deterministic-path likelihoods or changing the supervised context distribution, we estimate the marginal response-policy gradients by approximating posterior reveal-edge occupancies under a controllable scoring-state budget.

\vspace{-0.2cm}
\paragraph{Reinforcement learning for MDMs.}
Recent work has begun adapting RL and preference optimization to diffusion language models. d1~\citep{zhao2026d} introduces a post-training recipe for diffusion LMs, including a critic-free GRPO-style RL method for reasoning. SPG~\citep{wang2025spg} addresses the intractability of dLLM likelihoods by constructing sandwiched policy-gradient estimators. Other methods formulate denoising as a sequential decision process, including MDPO-style denoising-trajectory optimization~\citep{he2025mdpo}, learned unmasking policies~\citep{jazbec2025learning}, dUltra~\citep{chen2025dultra}, which introduces an additional unmasking planner optimized with GRPO for faster parallel decoding, and TraceRL~\cite{wang2025revolutionizing}, introducing an explicit value model for sampled trajectories. In contrast, \methodname does not introduce a separate reveal-policy head or rely on a pre-defined masking objective or a single sampled reveal path. We discuss additional related work in Appendix~\ref{appx:related-work}.
\vspace{-0.2cm}

\section{Conclusion}
\label{sec:conclusion}
\vspace{-0.2cm}
We introduced \methodname, a principled framework for reinforcement learning of masked diffusion language models with endogenous reveal orders. By combining a reveal-state DAG with budgeted frontier estimation, it connects the marginal-policy objective to a practical token-score surrogate.

\vspace{-0.2cm}
\paragraph{Limitations and future work.}
FRM approximates the reveal-order marginal under a finite computation budget. For the confidence-based sampler, local next-reveal probabilities can be computed exactly, but finite frontier coverage, approximate continuation weights, macro reveals, and the token-score surrogate still limit marginal-gradient accuracy. Its effectiveness therefore depends on the rollout sampler and the allocation of scoring evaluations. A broader study of reveal heuristics, macro-reveal selection, and frontier allocation strategies would further clarify when marginal edge coverage is most beneficial. Finally, improving the underlying MDM architecture and memory efficiency remains important for scaling marginal-policy optimization to longer responses and larger models.

\begin{ack}
This work was funded in part by the National Science Foundation (NSF) Awards IIS-2416835, DARPA I2O CARVE, NIST 60NANB26D035, Cisco and Amazon. Any opinions, findings and conclusions or recommendations expressed in this material are those of the authors and do not necessarily reflect the views of the sponsors.

The authors declare no competing interests.
\end{ack}

\bibliography{ref}
\bibliographystyle{plain}

\clearpage
\appendix

\section{Additional Related Work}
\label{appx:related-work}

\paragraph{Diffusion Language Models.}
Continuous diffusion models for image generation established the foundation of iterative refinement via noise reversal~\cite{song2020score,ho2020denoising}, while discrete diffusion models adapt these stochastic processes to categorical data domains~\cite{austin2021structured,shi2024simplified,campbell2024generative}. This progression demonstrates the effectiveness of diffusion mechanisms across different data modalities.
Continuous diffusion models in the text embedding space address the challenge of discrete text generation by mapping tokens to a continuous latent space for diffusion processes~\cite{li2022diffusion,gong2022diffuseq,han2023ssd,sahoo2025diffusion}. By operating on embeddings, these methods utilize the well-established mathematical framework of continuous diffusion while avoiding direct operations on discrete variables.
Masked diffusion language models (MDLMs) operate directly on discrete tokens by replacing the Gaussian noise process with a corruption process based on masking~\cite{lou2024discrete,zheng2023reparameterized,sahoo2024simple,nie2026large,ye2025dream}. This approach simplifies the learning objective to recovering masked information, effectively modeling the joint probability distribution of the text.
Combining continuous and discrete diffusion for text generation seeks to integrate the optimization benefits of continuous spaces with the semantic integrity of discrete modeling~\cite{von2025generalized,pynadath2025candi}. Such hybrid methodologies aim to enhance generation fidelity and convergence stability.
Block-diffusion interpolates autoregressive and masked diffusion generation, allowing for the simultaneous prediction of token blocks while maintaining sequential dependencies between blocks~\cite{arriola2025block,cheng2025sdar}. Another way to view block-diffusion is semi-autoregressive generation, unifying autoregressive and diffusion generation to bridge the gap between purely sequential and fully parallel decoding strategies~\cite{fathi2025unifying,liu2025sequential}.
Another notable trend in diffusion LMs is to generate flexible length, which is more intuitive and implies more powerful abilities of diffusion models by decoupling generation from a predetermined sequence length~\cite{havasi2025edit,kim2025any}. Enabling variable-length outputs allows these models to better accommodate the dynamic nature of natural language sequences.

\paragraph{Inference-Time Reveal-Order Search and Planning.}
\label{app:related_inference_planning}
A complementary line of work improves masked diffusion LM generation by modifying the inference-time unmasking path. Several methods use confidence or uncertainty to accelerate decoding or select better revealing trajectories. LocalLeap~\cite{kong2025localleap} exploits local determinism around high-confidence anchors for adaptive parallel decoding, while FreeDave~\cite{wu2025freedave} uses draft-and-verification to reduce the number of diffusion forwards. Other methods treat decoding more explicitly as a path-selection problem: Lookahead Unmasking~\cite{lee2025lookahead} evaluates candidate unmasking paths with uncertainty-based verification, Denoising Entropy~\cite{chen2025denoisingentropy} scores generation paths by cumulative predictive uncertainty, and Path Planning~\cite{peng2025pathplanning} separates planning from denoising at each step.

Recent work also explores broader search over both orders and token values. Order-Token Search~\cite{shen2026ordertokensearch} performs beam-style search jointly over generation order and token assignments, and MEDAL~\cite{huang2026medal} applies Monte Carlo tree search to explore promising denoising trajectories. These inference-time methods are closely related to our reveal-state DAG perspective: they also recognize that generation quality depends on the path through partially revealed states. However, their goal is to choose or verify a good decoding trajectory at test time. In contrast, \methodname uses reveal-order structure during RL training to approximate the marginal response-policy gradient. FRM does not search for a single best inference path; it constructs a budgeted set of posterior-relevant reveal edges for policy-gradient estimation.

\section{Theory}
\label{appx:theory}

\subsection{Proof of \autoref{thm:rb_policy_gradient}}
\label{appx:proof-rb-policy-gradient}

We prove \autoref{thm:rb_policy_gradient} in three steps. Throughout this section, expectations are taken over
\[
    c\sim \rho,
    \qquad
    (x,\sigma)\sim P_\theta(\cdot,\cdot\mid c),
\]
unless otherwise specified. We assume that the reward \(R(c,x)\) is fixed with respect to \(\theta\), and that differentiation can be exchanged with the finite sum over responses and reveal orders. The marginal response policy is
\[
    \pi_\theta(x\mid c)
    =
    \sum_\sigma P_\theta(x,\sigma\mid c).
\]

\paragraph{Value depends only on the marginal response policy.}
Since the reward depends only on the context \(c\) and completed response \(x\), not on the latent reveal order \(\sigma\), we have
\begin{align}
    J(\theta)
    &=
    \mathbb E_{c\sim\rho}
    \sum_{x,\sigma}
    P_\theta(x,\sigma\mid c)R(c,x) \\
    &=
    \mathbb E_{c\sim\rho}
    \sum_x
    \left(
        \sum_\sigma P_\theta(x,\sigma\mid c)
    \right)
    R(c,x) \\
    &=
    \mathbb E_{c\sim\rho}
    \sum_x
    \pi_\theta(x\mid c)R(c,x).
\end{align}
Thus, for sequence-level rewards, the conditional reveal-order distribution
\(P_\theta(\sigma\mid c,x)\) describes the latent execution trace. The reveal process affects how the response is generated, but the objective depends on it only through the induced marginal response policy \(\pi_\theta(x\mid c)\).

\paragraph{Marginal score as a Rao--Blackwellized path score.}
For any \(c,x\) with \(\pi_\theta(x\mid c)>0\),
\begin{align}
    \mathbb E_{\sigma\sim P_\theta(\cdot\mid c,x)}
    \left[
        \nabla_\theta \log P_\theta(x,\sigma\mid c)
    \right]
    &=
    \sum_\sigma
    P_\theta(\sigma\mid c,x)
    \frac{
        \nabla_\theta P_\theta(x,\sigma\mid c)
    }{
        P_\theta(x,\sigma\mid c)
    } \\
    &=
    \sum_\sigma
    \frac{
        P_\theta(x,\sigma\mid c)
    }{
        \pi_\theta(x\mid c)
    }
    \frac{
        \nabla_\theta P_\theta(x,\sigma\mid c)
    }{
        P_\theta(x,\sigma\mid c)
    } \\
    &=
    \frac{
        \sum_\sigma \nabla_\theta P_\theta(x,\sigma\mid c)
    }{
        \pi_\theta(x\mid c)
    } \\
    &=
    \frac{
        \nabla_\theta \pi_\theta(x\mid c)
    }{
        \pi_\theta(x\mid c)
    } \\
    &=
    \nabla_\theta \log \pi_\theta(x\mid c).
\end{align}
Therefore, the marginal response-policy score is the conditional expectation of the joint path score:
\[
    \nabla_\theta \log \pi_\theta(x\mid c)
    =
    \mathbb E
    \left[
        \nabla_\theta \log P_\theta(x,\sigma\mid c)
        \mid x,c
    \right].
\]

Now define
\[
    g_{\mathrm{traj}}(x,\sigma,c)
    =
    R(c,x)\nabla_\theta\log P_\theta(x,\sigma\mid c),
\]
and
\[
    g_{\mathrm{marg}}(x,c)
    =
    R(c,x)\nabla_\theta\log \pi_\theta(x\mid c).
\]
Since \(R(c,x)\) is measurable with respect to \((c,x)\), the previous identity implies
\[
    g_{\mathrm{marg}}(x,c)
    =
    \mathbb E
    \left[
        g_{\mathrm{traj}}(x,\sigma,c)
        \mid x,c
    \right].
\]
This proves the Rao--Blackwell relation in \autoref{thm:rb_policy_gradient}.

\paragraph{Unbiasedness.}
The trajectory estimator is unbiased:
\begin{align}
    \mathbb E[g_{\mathrm{traj}}]
    &=
    \mathbb E_{c\sim\rho}
    \sum_{x,\sigma}
    P_\theta(x,\sigma\mid c)
    R(c,x)
    \nabla_\theta \log P_\theta(x,\sigma\mid c) \\
    &=
    \mathbb E_{c\sim\rho}
    \sum_{x,\sigma}
    R(c,x)
    \nabla_\theta P_\theta(x,\sigma\mid c) \\
    &=
    \nabla_\theta
    \mathbb E_{c\sim\rho}
    \sum_{x,\sigma}
    P_\theta(x,\sigma\mid c)R(c,x) \\
    &=
    \nabla_\theta J(\theta).
\end{align}
Similarly, using the marginal response policy,
\begin{align}
    \mathbb E[g_{\mathrm{marg}}]
    &=
    \mathbb E_{c\sim\rho}
    \sum_x
    \pi_\theta(x\mid c)
    R(c,x)
    \nabla_\theta\log \pi_\theta(x\mid c) \\
    &=
    \mathbb E_{c\sim\rho}
    \sum_x
    R(c,x)
    \nabla_\theta \pi_\theta(x\mid c) \\
    &=
    \nabla_\theta
    \mathbb E_{c\sim\rho}
    \sum_x
    \pi_\theta(x\mid c)R(c,x) \\
    &=
    \nabla_\theta J(\theta).
\end{align}
Thus,
\[
    \mathbb E[g_{\mathrm{traj}}]
    =
    \mathbb E[g_{\mathrm{marg}}]
    =
    \nabla_\theta J(\theta).
\]

\paragraph{Variance dominance.}
Since
\[
    g_{\mathrm{marg}}
    =
    \mathbb E[g_{\mathrm{traj}}\mid x,c],
\]
the law of total covariance gives
\[
    \operatorname{Cov}(g_{\mathrm{traj}})
    =
    \operatorname{Cov}
    \left(
        \mathbb E[g_{\mathrm{traj}}\mid x,c]
    \right)
    +
    \mathbb E
    \left[
        \operatorname{Cov}(g_{\mathrm{traj}}\mid x,c)
    \right].
\]
Using the Rao--Blackwell relation,
\[
    \operatorname{Cov}(g_{\mathrm{traj}})
    =
    \operatorname{Cov}(g_{\mathrm{marg}})
    +
    \mathbb E
    \left[
        \operatorname{Cov}(g_{\mathrm{traj}}\mid x,c)
    \right].
\]
The conditional covariance term is positive semidefinite, so
\[
    \operatorname{Cov}(g_{\mathrm{marg}})
    \preceq
    \operatorname{Cov}(g_{\mathrm{traj}}).
\]
Equivalently, for any test direction \(u\),
\[
    \operatorname{Var}(u^\top g_{\mathrm{traj}})
    -
    \operatorname{Var}(u^\top g_{\mathrm{marg}})
    =
    \mathbb E
    \left[
        \operatorname{Var}(u^\top g_{\mathrm{traj}}\mid x,c)
    \right]
    \ge 0.
\]
Since
\[
    g_{\mathrm{traj}}
    =
    R(c,x)
    \nabla_\theta \log P_\theta(x,\sigma\mid c),
\]
this conditional variance can also be written as
\[
    \mathbb E
    \left[
        R(c,x)^2
        \operatorname{Var}
        \left(
            u^\top
            \nabla_\theta \log P_\theta(x,\sigma\mid c)
            \mid x,c
        \right)
    \right].
\]
Therefore, equality holds if and only if
\[
    \operatorname{Var}(g_{\mathrm{traj}}\mid x,c)=0
\]
almost surely. When \(R(c,x)\neq 0\), this is equivalent to
\[
    \nabla_\theta \log P_\theta(x,\sigma\mid c)
\]
being almost surely constant over reveal orders \(\sigma\) conditional on \(x,c\).

\section{Endogenous Reveal Probabilities for Confidence-Based Decoding}
\label{app:local_order}
\label{app:scorelaw}

This section describes LLaDA's low-confidence remasking rule~\cite{nie2026large} with the temperature-controlled token sampling used in its released implementation.\footnote{\url{https://github.com/ML-GSAI/LLaDA/blob/main/generate.py}} We derive the probabilities of revealing specified response tokens in a general \(k\)-token update and its one-token special case, then describe their computation and cost.

\subsection{Confidence-Based Decoding}

Given a context \(c\), a response \(x\), and a revealed set \(S\), the partial response \(\tilde x_S\) contains \(x^i\) for \(i\in S\) and masks at the remaining response positions. Let \(\mathcal I(S)\) be the masked positions eligible for the next update: those in the active block for blockwise decoding, or all masked response positions otherwise.

One model evaluation gives logits \(\mathbf a_{\theta,j}=\operatorname{logits}_\theta(c,\tilde x_S)_j\) over vocabulary \(\mathcal V\). At temperature \(\tau>0\), candidates at eligible positions are drawn independently:
\begin{equation}
    \label{eq:local-cdf-token-laws}
    p_{\theta,j}(v)=\operatorname{softmax}(\mathbf a_{\theta,j}/\tau)_v,
    \qquad Z_j\sim\mathrm{Cat}(\cdot;p_{\theta,j}).
\end{equation}
The confidence assigned to \(Z_j\) is \(\operatorname{softmax}(\mathbf a_{\theta,j})_{Z_j}\), its probability under the untempered model output. Temperature therefore changes which tokens are sampled, but not how their confidence is measured. At \(\tau=0\), the decoder chooses an argmax token deterministically.

Rank the candidates by decreasing confidence, with lower position indices winning ties. Let \(\operatorname{Top}_k(Z)\) be the set of the first \(k\) positions. For a scheduled reveal count \(1\le k\le|\mathcal I(S)|\), the decoder reveals the positions \(\Delta S\) and updates the state as
\begin{equation}
    \label{eq:confidence-decoder-selection}
    \Delta S=\operatorname{Top}_k(Z),\qquad S'=S\cup \Delta S.
\end{equation}
Using the predictions at \(\tilde x_S\), the decoder fills each selected position \(j\in\Delta S\) with its sampled token \(Z_j\), leaving the remaining eligible positions masked.

\subsection{Target-Conditioned Reveal Probabilities}

Revealing specified response tokens requires both proposing those tokens and selecting their positions. Once their candidate values are fixed, selection depends on the remaining sampled candidates. The confidence-based rule turns this event into independent confidence comparisons, whose probabilities are obtained from cumulative distribution functions (CDFs).

\paragraph{\(k\)-token reveals.}
For a queried reveal set \(\Delta S\subseteq\mathcal I(S)\) with \(|\Delta S|=k\), we condition on its candidates matching the response tokens, \(Z_{\Delta S}=x_{\Delta S}\). The \emph{competing candidates} are the sampled tokens \(Z_j\) at the remaining eligible positions \(j\in\mathcal I(S)\setminus\Delta S\). This conditioning fixes the target candidates' confidences, while the competing candidates remain random and independent.

Let \(i_{\Delta S}\) be the position of the lowest-ranked target token in \(\Delta S\). The queried set is selected precisely when every competing candidate ranks below this token. We therefore compute the probability of each such comparison and multiply the results.

At each competing position \(j\), confidence values are fixed by the logits, while the candidate \(Z_j\) is random. The strict and non-strict CDFs of its confidence at threshold \(u\) are
\begin{equation}
    \label{eq:local-cdf-definition}
    \begin{aligned}
    F_{\theta,j}^{<}(u)
        &=\sum_{v\in\mathcal V}p_{\theta,j}(v)
          \mathbf 1\!\left\{\operatorname{softmax}(\mathbf a_{\theta,j})_v<u\right\},\\
    F_{\theta,j}^{\le}(u)
        &=\sum_{v\in\mathcal V}p_{\theta,j}(v)
          \mathbf 1\!\left\{\operatorname{softmax}(\mathbf a_{\theta,j})_v\le u\right\}.
    \end{aligned}
\end{equation}
Each sum weights the possible candidate tokens by their sampling probabilities. For example, \(F_{\theta,j}^{\le}(u)\) is the probability that the sampled token at position \(j\) has confidence at most \(u\).

The conditional selection probability and joint transition probability are
\begin{equation}
    \label{eq:local-cdf-macro-event}
    \begin{aligned}
    r_\theta^x(\Delta S\mid c,\tilde x_S)
        &=\Pr\!\left(\operatorname{Top}_k(Z)=\Delta S\mid c,\tilde x_S,Z_{\Delta S}=x_{\Delta S}\right),\\
    \psi_\theta(S,\Delta S;c,x)
        &=\left(\prod_{i\in \Delta S}p_{\theta,i}(x^i)\right)
          r_\theta^x(\Delta S\mid c,\tilde x_S).
    \end{aligned}
\end{equation}
The second line multiplies the target-token sampling probabilities by the conditional selection probability. Evaluating every competing candidate's CDF at the confidence of \(x^{i_{\Delta S}}\) gives
\begin{equation}
    \label{eq:local-cdf-set-product}
    \begin{aligned}
    r_\theta^x(\Delta S\mid c,\tilde x_S)
      ={}&\prod_{\substack{j\in\mathcal I(S)\setminus \Delta S\\j<i_{\Delta S}}}
        F_{\theta,j}^{<}\!\left(\operatorname{softmax}(\mathbf a_{\theta,i_{\Delta S}})_{x^{i_{\Delta S}}}\right)\\
      &\quad\cdot\prod_{\substack{j\in\mathcal I(S)\setminus \Delta S\\j>i_{\Delta S}}}
        F_{\theta,j}^{\le}\!\left(\operatorname{softmax}(\mathbf a_{\theta,i_{\Delta S}})_{x^{i_{\Delta S}}}\right).
    \end{aligned}
\end{equation}
If \(\Delta S=\mathcal I(S)\), there are no competitors and the selection probability is one. These conditional probabilities assume that all queried target tokens have positive sampling probability; otherwise the joint transition probability is zero.

\paragraph{One-token reveals.}
For \(k=1\), set \(\Delta S=\{i\}\) and let \(I\) denote the highest-ranked position. Writing the singleton query by its position index gives
\begin{equation}
    \label{eq:local-cdf-product}
    \begin{aligned}
    r_\theta^x(i\mid c,\tilde x_S)
    &=\Pr(I=i\mid c,\tilde x_S,Z_i=x^i)\\
    &=\prod_{\substack{j\in\mathcal I(S)\\j<i}}
       F_{\theta,j}^{<}\!\left(\operatorname{softmax}(\mathbf a_{\theta,i})_{x^i}\right)
     \prod_{\substack{j\in\mathcal I(S)\\j>i}}
       F_{\theta,j}^{\le}\!\left(\operatorname{softmax}(\mathbf a_{\theta,i})_{x^i}\right).
    \end{aligned}
\end{equation}
This is the one-token selection probability in Eq.~\ref{eq:local-cdf-selection}. Multiplying by the probability of sampling \(x^i\) gives the corresponding edge factor,
\begin{equation}
    \label{eq:local-cdf-edge}
    \psi_\theta(S,i;c,x)
    =p_{\theta,i}(x^i)\,r_\theta^x(i\mid c,\tilde x_S).
\end{equation}
Each \(r_\theta^x(i\mid c,\tilde x_S)\) conditions on a different event, \(Z_i=x^i\). These probabilities therefore need not sum to one over positions, and the edge factor uses them without cross-position normalization.

\begin{takeawaybox}
\begin{remark}[Parallel decoding]
For parallel decoding that reveals \(k>1\) tokens per update, the corresponding transition is the \(k\)-reveal kernel in Eq.~\ref{eq:local-cdf-macro-event}, with all selected tokens predicted at the same state. The current FEROM formulation uses one-token transitions; extending frontier marginalization directly to \(k\)-reveal transitions is left to future work.
\end{remark}
\end{takeawaybox}

\subsection{Computation and Cost}
At each eligible position, sort vocabulary tokens by increasing confidence and store cumulative sums of their sampling probabilities. A binary search then gives the total probability below, or at most, a supplied confidence threshold. For a reveal-set query, use the confidence of its lowest-ranked target as this threshold at each competing position. Summing the log-CDF values evaluates the product stably. Algorithm~\ref{alg:local-cdf} summarizes the calculation.

Let \(m=|\mathcal I(S)|\) and \(V=|\mathcal V|\). Building the sorted tables takes \(O(mV\log V)\) time and \(O(mV)\) storage. Once built, a query costs \(O(|\Delta S|+(m-|\Delta S|)\log V)\): find the lowest-ranked target, then look up the competitors' CDFs. Computing all \(m\) singleton queries therefore takes \(O(mV\log V+m^2\log V)\) total time. If their CDF factors are cached, storage becomes \(O(mV+m^2)\). These calculations reuse the existing logits and require no additional model evaluation; processing positions in chunks reduces peak table memory.

\begin{algorithm}[H]
\caption{Computing Target-Conditioned Reveal-Set Probabilities}
\label{alg:local-cdf}
\label{alg:scorelaw}
\begin{algorithmic}[1]
\Require Current-state logits, response \(x\), eligible positions \(\mathcal I(S)\), temperature \(\tau\), and nonempty reveal sets whose target tokens have positive sampling probability
\For{\(j\in\mathcal I(S)\)}
    \State Compute token-sampling probabilities \(p_{\theta,j}\) and confidences \(\operatorname{softmax}(\mathbf a_{\theta,j})\)
    \State Sort tokens by increasing confidence and store cumulative sampling probabilities
\EndFor
\For{each queried set \(\Delta S\subseteq\mathcal I(S)\)}
    \State Find the lowest-ranked target position \(i_{\Delta S}\)
    \State Look up the competing positions' CDFs at the confidence of \(x^{i_{\Delta S}}\)
    \State Compute conditional selection probability \(r_\theta^x(\Delta S\mid c,\tilde x_S)\) by Eq.~\ref{eq:local-cdf-set-product}
    \State Compute joint transition probability \(\psi_\theta(S,\Delta S;c,x)\) by Eq.~\ref{eq:local-cdf-macro-event}
\EndFor
\State \Return conditional selection and joint transition probabilities for the queried sets
\end{algorithmic}
\end{algorithm}

\section{Computing Posterior Occupancies on the Reveal-State DAG}
\label{app:exact_dag_dp}

This section derives the backward recursion for completion mass and the forward propagation of node and edge occupancies on the reveal-state DAG. These recursions relate the local edge factors to the response-conditioned next-reveal posterior and marginal-score decomposition used in \S\ref{sec:reveal_state_dag}.

\paragraph{Setup.}
We consider a context \(c\) and a sequence \(x\) for which \(\pi_\theta(x\mid c)>0\). Let \(S_0\subseteq[n]\) be the initial revealed set, with any initially fixed tokens included in the conditioning context \(c\); ordinary response generation uses \(S_0=\emptyset\). A reveal order \(\sigma=(\sigma_1,\ldots,\sigma_{n-|S_0|})\) is a permutation of the initially masked positions \([n]\setminus S_0\). Define the revealed set after \(t\) reveal steps as
\[
    S_t(\sigma)
    =
    S_0\cup\{\sigma_1,\ldots,\sigma_t\}.
\]
Thus, \(S_0(\sigma)=S_0\), and \(S_{n-|S_0|}(\sigma)=[n]\).

For a revealed set \(S\) and an eligible next position \(i\in\mathcal I(S)\), define the local edge factor
\[
    \psi_\theta(S,i;c,x)
    \triangleq
    r_\theta^x(i\mid c,\tilde x_S)\,
    p_\theta(x^i\mid c,\tilde x_S).
\]
The first term is the probability of selecting \(i\) conditional on its proposal being \(x^i\), with competitors marginalized. The second is the probability of proposing that token under the actual rollout distribution. Set \(\psi_\theta(S,i;c,x)=0\) for ineligible positions, so the following sums also cover restricted schedules such as blockwise decoding. For a full reveal order \(\sigma\), the joint path probability of generating \(x\) through \(\sigma\) is
\[
    P_\theta(x,\sigma\mid c)
    =
    \prod_{t=0}^{n-|S_0|-1}
    \psi_\theta(S_t(\sigma),\sigma_{t+1};c,x).
\]

\paragraph{Completion mass.}
For any revealed set \(S\supseteq S_0\), define the completion mass
\[
    \mathcal Z_\theta(S;c,x)
\]
as the total probability mass of all reveal continuations that complete the response \(x\) from state \(S\). Equivalently,
\[
    \mathcal Z_\theta(S;c,x)
    =
    \sum_{\text{reveal orders from }S\text{ to }[n]}
    \prod_{\text{edges }(S',i)\text{ on the continuation}}
    \psi_\theta(S',i;c,x).
\]
This is a path-sum normalizer, not a learned value function. It satisfies the backward recursion
\[
    \mathcal Z_\theta(S;c,x)
    =
    \sum_{i\notin S}
    \psi_\theta(S,i;c,x)\,
    \mathcal Z_\theta(S\cup\{i\};c,x),
    \qquad
    \mathcal Z_\theta([n];c,x)=1.
\]
The marginal response policy is the completion mass at the initial state:
\[
    \pi_\theta(x\mid c)
    =
    \sum_\sigma P_\theta(x,\sigma\mid c)
    =
    \mathcal Z_\theta(S_0;c,x).
\]

\paragraph{Response-conditioned next-reveal posterior.}
The target-conditioned selection probability \(r_\theta^x(i\mid c,\tilde x_S)\) conditions on one proposed token, not on producing the entire response. The response-conditioned next-reveal posterior instead asks: among all paths that generate \(x\) and have reached \(S\), how likely is the next edge to reveal position \(i\)?

For any nonterminal state \(S\) reached with positive probability under \(P_\theta(\sigma\mid c,x)\), this posterior is
\[
    \eta_\theta(i\mid S,c,x)
    =
    \Pr_\theta
    \left[
        \sigma_{|S|-|S_0|+1}=i
        \mid
        S_{|S|-|S_0|}(\sigma)=S,\;c,\;x
    \right],
\]
and it has the closed form
\[
    \eta_\theta(i\mid S,c,x)
    =
    \frac{
        \psi_\theta(S,i;c,x)\,
        \mathcal Z_\theta(S\cup\{i\};c,x)
    }{
        \mathcal Z_\theta(S;c,x)
    }.
\]
Equivalently,
\[
    \eta_\theta(i\mid S,c,x)
    \propto
    r_\theta^x(i\mid c,\tilde x_S)\,
    p_\theta(x^i\mid c,\tilde x_S)\,
    \mathcal Z_\theta(S\cup\{i\};c,x).
\]
Thus the posterior combines the conditional selection probability, the likelihood of proposing the target token, and the future completion mass. Local conditioning alone does not recover this posterior. Values of \(\eta_\theta\) at states with zero posterior occupancy do not contribute to the marginal score.

\paragraph{Node and edge occupancies.}
The response-conditioned posterior over reveal trajectories induces occupancies on the reveal-state DAG. The oracle node occupancy is
\[
    \nu_\theta^\star(S;c,x)
    =
    \Pr_{\sigma\sim P_\theta(\sigma\mid c,x)}
    \left[
        S_{|S|-|S_0|}(\sigma)=S
    \right],
\]
for \(S\supseteq S_0\), and \(\nu_\theta^\star(S;c,x)=0\) otherwise. The initial occupancy is
\[
    \nu_\theta^\star(S_0;c,x)=1.
\]
For an edge \(e=(S,i)\), the oracle edge occupancy is
\[
    \mu_{\theta,(S,i)}^\star(c,x)
    =
    \Pr_{\sigma\sim P_\theta(\sigma\mid c,x)}
    \left[
        S_{|S|-|S_0|}(\sigma)=S,\;
        \sigma_{|S|-|S_0|+1}=i
    \right].
\]
Using the response-conditioned next-reveal posterior, the edge occupancy factorizes as
\[
    \mu_{\theta,(S,i)}^\star(c,x)
    =
    \nu_\theta^\star(S;c,x)\,
    \eta_\theta(i\mid S,c,x).
\]
Node occupancies can also be propagated forward through incoming edges. For \(S\neq S_0\),
\[
    \nu_\theta^\star(S;c,x)
    =
    \sum_{j\in S\setminus S_0}
    \mu_{\theta,(S\setminus\{j\},j)}^\star(c,x).
\]
This recurrence is the exact counterpart of the approximate node-occupancy propagation used by FRM.

\paragraph{Edge-occupancy form of the marginal score.}
Assume the relevant edge factors are differentiable and positive on the support of the posterior. The marginal score can be written as an edge-occupancy weighted sum. Starting from
\[
    \pi_\theta(x\mid c)
    =
    \sum_\sigma
    P_\theta(x,\sigma\mid c),
\]
we have
\begin{align}
    \nabla_\theta \log \pi_\theta(x\mid c)
    &=
    \sum_\sigma
    P_\theta(\sigma\mid c,x)
    \nabla_\theta \log P_\theta(x,\sigma\mid c) \\
    &=
    \sum_\sigma
    P_\theta(\sigma\mid c,x)
    \sum_{t=0}^{n-|S_0|-1}
    \nabla_\theta
    \log
    \psi_\theta(S_t(\sigma),\sigma_{t+1};c,x) \\
    &=
    \sum_{e=(S,i)}
    \mu_{\theta,e}^\star(c,x)
    \nabla_\theta
    \log
    \psi_\theta(S,i;c,x) \\
    &=
    \sum_{e=(S,i)}
    \mu_{\theta,e}^\star(c,x)
    \nabla_\theta
    \log
    \left[
        r_\theta^x(i\mid c,\tilde x_S)
        p_\theta(x^i\mid c,\tilde x_S)
    \right].
\end{align}
This is the exact reveal-state DAG form of the Rao--Blackwellized marginal policy score.
The posterior occupancies here are coefficients from differentiating the original path sum (Fisher's identity). One may detach them when evaluating this weighted score; no extra derivative of occupancy is required. The score of \(r_\theta^x\), however, is part of the differentiated edge factor.

\paragraph{Relation to the token-score surrogate.}
The practical estimator in the main paper uses the reveal process to estimate posterior edge occupancies, but applies gradients only to the token-likelihood component. For edge \(e=(S,i)\), define
\[
    h_e
    =
    \nabla_\theta
    \log p_\theta(x^i\mid c,\tilde x_S).
\]
The oracle token-score target is
\[
    g^\star_{\mathrm{tok}}(c,x)
    =
    \sum_e
    \mu_{\theta,e}^\star(c,x) h_e.
\]
This omits the reveal-kernel score, so it need not equal the full marginal score even with exact occupancies. Rather, it is the token-score component of the edge-occupancy decomposition, and it is the target approximated by the budgeted FRM estimator in the main text.

To separate the errors without assuming that macro reveals share the one-token edge basis, let \(\widehat g_B\) be the practical token-score estimate and \(k_e=\nabla_\theta\log r_\theta^x(i\mid c,\tilde x_S)\). Then
\[
    \widehat g_B-\nabla_\theta\log\pi_\theta(x\mid c)
    =\bigl(\widehat g_B-g^\star_{\mathrm{tok}}\bigr)
    -\sum_e\mu_{\theta,e}^\star k_e.
\]
The first term includes finite-frontier coverage, approximate continuation weighting, and the shared-parent approximation in macro reveals. The one-token local probabilities are computed exactly, but the macro proposals and response-conditioned occupancies remain approximate. The second term is the omitted reveal score. This is an error decomposition, not a bound: exact local probabilities alone do not guarantee a small total error. Theorem~\ref{thm:rb_policy_gradient} concerns exact complete scores and exact posterior averaging, not variance dominance of \(\widehat g_B\).

\section{Example of Oracle Marginal Distribution}
\label{appx:example-oracle-marginal}

\begin{figure}
    \centering
    \includegraphics[width=1.0\linewidth]{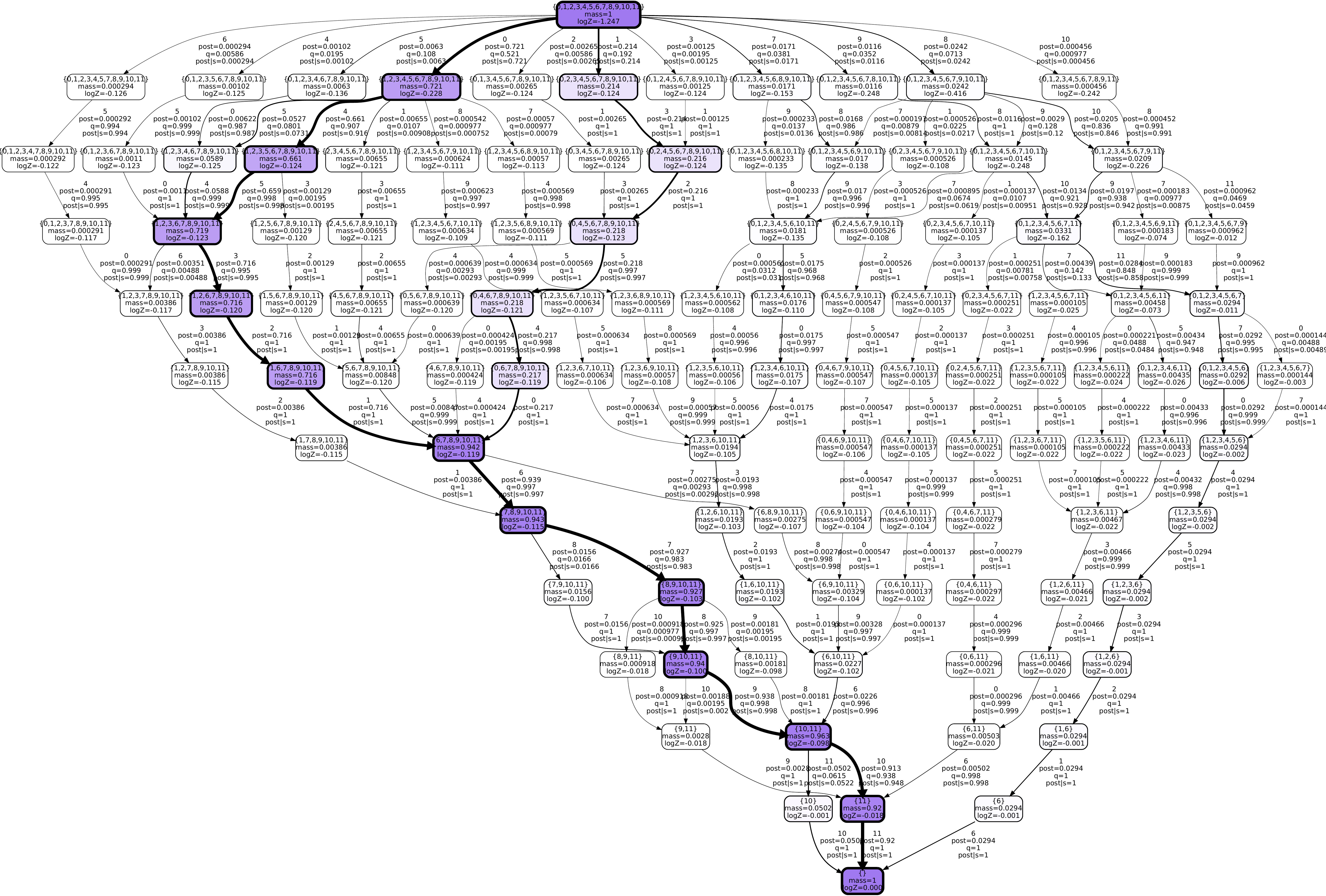}
    \caption{Reveal-state DAG for a 12-position block from an AceCoder completion.}
    \label{fig:oracle-poset}
\end{figure}

Figure~\ref{fig:oracle-poset} shows a reveal-state DAG for a 12-position block from an AceCoder completion. The distribution is concentrated but supports several plausible reveal trajectories, motivating marginalization over reveal orders.

\Needspace{14\baselineskip}
\section{Frontier Reveal Marginalization Details}
\label{app:frm_variants}
This appendix specifies how FRM constructs macro reveals, allocates state evaluations across continuations, and combines their token scores. It uses the local planning distribution in Eq.~\ref{eq:frm-local-target}; numerical settings for the experiments are collected in Appendix~\ref{app:experimental-protocols}.

\subsection{Macro-Reveal Construction}
\label{app:frm-macro}
\paragraph{Budget-aware reveal size.}
Let \(b_S\ge1\) be the number of state evaluations assigned to a continuation, including the current evaluation. The nominal reveal count is
\begin{equation}
    \label{eq:macro-horizon}
    k_0(S)=\left\lceil\frac{|\mathcal I(S)|}{b_S}\right\rceil.
\end{equation}
This horizon spreads token coverage across the available states. For example, two state evaluations can cover 32 positions with two 16-token macro reveals. Each continuation updates its horizon as positions and evaluations are consumed.

\paragraph{Conditional confidence rankings.}
After selecting a leading position \(i\), FRM forms a macro reveal from positions likely to follow it in the same parent prediction. For a supported leading position, we condition on both \(Z_i=x^i\) and \(I=i\). Each competing candidate is drawn independently from its token distribution restricted to candidates that rank below this target, using the CDFs and tie convention in Appendix~\ref{app:local_order}. The first \(k\) positions of the resulting confidence ranking form the conditional prefix; for \(k=1\), it contains only \(i\).

For an unsplit continuation, the leading position follows \(p_{\mathrm{loc},\theta}\). Discarding the sampled prefix's order gives a macro-set with distribution
\begin{equation}
    \label{eq:conditional-prefix-macro-proposal}
    \begin{aligned}
    q_{k,\theta}(\Delta S\mid S,c,x)
    ={}&\sum_{i\in\Delta S}
        p_{\mathrm{loc},\theta}(i\mid S,c,x)\\
       &\quad\cdot\Pr_\theta\!\left(
        \operatorname{Top}_k(Z)=\Delta S
        \mid c,\tilde x_S,I=i,Z_i=x^i\right),
        \qquad |\Delta S|=k,
    \end{aligned}
\end{equation}
where the sum includes supported leading positions. We approximate the conditional set probabilities with empirical frequencies from sampled rankings, retaining \(p_{\mathrm{loc},\theta}\), renormalized over represented leading positions, as the mixture weights. The CDF calculation supplies the one-token selection probabilities, while the ranking samples supply the additional macro positions.

\paragraph{Coverage and concentration.}
The fixed-horizon rule uses \(k_0(S)\) directly. An optional size controller compares this count with neighboring counts using
\begin{equation}
    \label{eq:macro-count-controller}
    \operatorname{score}(k)
    =k\max_{\Delta S:\,|\Delta S|=k}
        \widehat q_k(\Delta S\mid S,c,x),
\end{equation}
where \(\widehat q_k\) is the finite-pool version of Eq.~\ref{eq:conditional-prefix-macro-proposal}. The factor \(k\) rewards token coverage; the largest set probability favors sizes for which candidate rankings agree. This criterion chooses the size, not the particular set: the macro-set is subsequently sampled. A search radius of one considers \(k_0-1,k_0,k_0+1\), clipped to the eligible range; ties favor the horizon and then the smaller count. Radius zero retains the fixed-horizon rule.

\subsection{Frontier Expansion and Contribution Weights}
\label{app:frm-frontier}
\paragraph{Continuations and allowed reveals.}
A frontier continuation carries its state \(S\), allowed leading positions, incoming weight \(\widehat\nu(S)\), and remaining state evaluations \(b_S\). The support restriction is suppressed in this notation. A residual continuation excludes leading choices already represented by other branches; \(p_{\mathrm{loc},\theta}\) is restricted and renormalized over that support. This restriction changes which position can lead, not which candidates participate in the parent confidence competition.

For blockwise decoding, each active block receives a share of the response-level budget and starts with incoming weight one. Root branching separates alternative leading reveals at a block's initial state. Local branching splits a continuation at an intermediate state \(S\), retaining the already-revealed positions in both branches. Both use the incoming weight \(\widehat\nu(S)\) in the contribution rule below, without duplicating earlier scored contributions. The split divides only the continuation's future evaluation budget; branch-count limits and secondary-budget caps remain block-level and do not reset at intermediate states.

\paragraph{Branch admission and budget allocation.}
Before sampling the scored macro-set, FRM considers separating the most likely allowed leading position \(i\) from its complement. With \(b_S\) state evaluations assigned, including the current evaluation, the future allocations satisfy
\begin{equation}
    \label{eq:frm-visit-split}
    b_{\mathrm{main}}+b_{\mathrm{res}}=b_S-1.
\end{equation}
For \(b_S>1\), write \(p_i=p_{\mathrm{loc},\theta}(i\mid S,c,x)\). The mass-proportional allocation rounds \((b_S-1)(1-p_i)\) to the nearest integer and raises it to at least one. This tentative secondary allocation is capped by the unused block-level secondary allowance and the per-split limit, yielding \(b_{\mathrm{res}}\). The per-split limit is a configured fraction of \(b_S-1\), rounded down with a minimum of one. The main continuation receives the remaining future evaluations, as in Eq.~\ref{eq:frm-visit-split}.

A split must respect the branch-count limit, give the residual its configured minimum number of evaluations, and leave at least one evaluation for the main child if it remains unfinished. The residual must have positive weight. Using an absolute threshold in expected state evaluations, admission additionally requires
\begin{equation}
    \label{eq:frm-branch-admission}
    \widehat\nu(S)(1-p_i)b_{\mathrm{res}}
    \ge \max\!\left\{\tau_{\mathrm{branch}},\lambda\,\widehat\nu(S)p_i b_{\mathrm{main}}\right\},
\end{equation}
where \(\tau_{\mathrm{branch}}\ge0\) is the absolute threshold and \(\lambda\ge0\) sets the required fraction of the main continuation's weight-times-budget value. These tests use the evaluated parent probabilities before drawing the scored prefix. If the split is rejected, all \(b_S-1\) future evaluations remain with one sampled continuation.

\paragraph{Weight accounting.}
Without a split, the leading position and its conditional prefix are sampled from the local proposal. Their selection probabilities are already represented by sampling frequency, so the macro-edge inherits \(\widehat\nu(S)\), as in Eq.~\ref{eq:frm-contribution-rule}. For an admitted split, the leading choice \(i\) is explicitly retained and only its conditional prefix is sampled. Its contribution and the complementary continuation's weight are
\begin{equation}
    \label{eq:frm-contribution-weight}
    \begin{aligned}
        \widehat\mu_e
            &=\widehat\nu(S)\,
              p_i,\\
        \widehat\nu(S)-\widehat\mu_e
            &=\widehat\nu(S)\,
              (1-p_i).
    \end{aligned}
\end{equation}
The residual samples its next leading position from the complementary support. These two contributions partition the incoming weight. Multiplying a sampled macro-edge by its proposal probability again would count that selection probability twice.

\paragraph{Expansion and merging.}
The child inherits the scored edge's coefficient by Eq.~\ref{eq:frm-child-contribution}; the residual remains a separate continuation at the parent. Pending continuations are prioritized by represented weight and, when a parent distribution is available, its spread:
\[
    \operatorname{priority}(S)
        =\frac{\widehat\nu(S)}{\sqrt{\sum_{j\in\mathcal I(S)}
            p_{\mathrm{loc},\theta}(j\mid S,c,x)^2}}.
\]
Residual continuations retain their original parent distribution and use remaining weight as priority. Continuations with the same state, allowed leading positions, and parent distribution can merge by adding their weights and remaining budgets. Each represented contribution is propagated once. Planning ends when the budget is exhausted or no unfinished continuation remains.

\subsection{Scoring and Training Conventions}
\label{app:frm-scoring}
\paragraph{Shared-parent token scores.}
A macro-edge records \((S,i,\Delta S)\), retaining the leading position for weight accounting. Every target token in \(\Delta S\) is scored at the same parent, as defined in \S\ref{sec:frm}; the next evaluated state includes the whole set. Macro reveals exchange intermediate-state resolution for token coverage. Appendix~\ref{app:exact_dag_dp} separates this approximation and finite frontier coverage from the omitted reveal-order score.

\paragraph{Aggregation and normalization.}
The token-score objective sums contributions with their raw coefficients from Eq.~\ref{eq:frm-contribution-rule}, without a softmax over selected edges. Coefficients are detached during differentiation. Training normalizes the loss by the configured generation length, using the same constant for all responses.

\paragraph{Scoring budget and runtime.}
The budget \(B\) counts scored state evaluations. We report the corresponding runtime and its breakdown in Appendix~\ref{app:timing-protocol}.

\section{Experimental Protocols}
\label{appx:training-details}
\label{app:experimental-protocols}
We summarize the settings for the main-table experiments, offline study, training ablations, and runtime profile. For confidence-based generation, the number of decoding steps equals the generation length.

\subsection{Main-Table Training}
\label{app:main-training}
We train LLaDA-8B-Instruct~\cite{nie2026large} with 4-bit QLoRA and bfloat16 mixed precision on eight NVIDIA H200 GPUs. The LoRA adapters~\cite{hu2022lora} use rank 128, scaling \(\alpha=64\), and dropout 0.05. Table~\ref{tab:training-common} gives the optimizer settings.
We use the training splits of \texttt{openai/gsm8k} (\texttt{main}) and \texttt{ankner/math-500} for GSM8K and MATH, respectively, and the ESPO-preprocessed hard subset of AceCoder-87K for coding. Training rollouts use confidence-based decoding with temperature 0.9, generation length 256, and block length 32. Each prompt has 16 sampled responses; updates average four unique prompts for GSM8K and MATH, and two to four for coding, using gradient accumulation.

GSM8K uses the question directly as its user prompt; MATH requests step-by-step reasoning and a boxed final answer. Both use final-answer correctness rewards without a separate format bonus. Coding follows the ESPO prompt and reward recipe: a system instruction requests a single final Python code block, the user prompt supplies the problem and an example test, and the reward combines a code-format score with the fraction of supplied tests passed. Execution credit requires correctly formatted, syntactically valid code.

\begin{table}[htb]
\centering
\caption{Optimizer settings for the main-table \methodname runs.}
\label{tab:training-common}
\small
\begin{tabular}{ll}
\toprule
 & Value \\
\midrule
Optimizer & AdamW, $\beta=(0.9, 0.99)$, weight decay 0.1 \\
LR schedule & Constant after warmup; warmup ratio \(10^{-4}\) \\
Learning rate & \(3\times10^{-6}\) \\
Gradient-norm clipping & 0.2 \\
KL coefficient & 0.0 \\
Clipping threshold $\epsilon$ & 0.5 \\
\bottomrule
\end{tabular}
\end{table}

The scoring allocation is two evaluated parent states per block for GSM8K and MATH, and 16 for AceCoder, matching the random-masking scoring budget of ESPO.

\subsection{Main-Table Evaluation}
\label{app:main-evaluation}
We use the JustGRPO-style evaluation pipeline~\cite{ni2026flexibility} with confidence-based decoding, block length 32, temperature zero, and no classifier-free guidance. Adapters are evaluated on an unquantized bfloat16 base model. We generate one completion per problem and report math accuracy and code pass@1. Baseline scores are taken from their source papers.

Each problem uses the model's chat template. GSM8K supplies the question, and MATH-500 additionally requests a boxed final answer. Code prompts request a complete Python function from the HumanEval specification or the MBPP problem, function stub, and example tests. We grade math answers with the task grader and code by functional correctness, evaluating 1,319 GSM8K, 500 MATH-500, 164 HumanEval, and 500 MBPP problems.

\subsection{Offline Gradient-Alignment Study}
\label{app:offline-protocol}

\paragraph{Data and continuation states.}
We generate eight 256-token responses per AceCoder prompt for 32 prompts, using LLaDA-8B-Instruct with confidence-based decoding, block length 32, and temperature 0.9. Rewards are supplied-test pass fractions. For prompts with reward variation, we uniformly select a four-response subset whose recomputed group-relative advantages are nonzero. We uniformly select a common block index that is complete before EOS in all source responses, taking each selected response's halfway state with 16 masked positions. This gives 56 continuation cases from 14 prompt groups. All methods share these states, targets, and advantages; zero-advantage prompts remain in aggregate averages.

\paragraph{Marginal reference and scoring.}
Dynamic programming over all subsets of the remaining positions gives the conditional marginal-gradient reference from the joint token/CDF transition probabilities. It includes both token-prediction and reveal-selection derivatives and is normalized by continuation length. Candidate scores include both components or only token derivatives, with detached coefficients. All gradients use the same LoRA initialization and random projection.

\paragraph{Budget and comparisons.}
Budgets range from 4 to 32 state evaluations. We compare FEROM with a single path using evenly spaced evaluations, the initial one-token steps of a path, left-to-right scoring, random masking, and two independent complete paths at the largest budget. Planning uses predictions only at evaluated states. Token-plus-order comparisons include methods with supported endogenous reveal transitions.

\paragraph{Metrics and parameter selection.}
Unweighted cosine averages case-wise gradient cosines. Advantage-weighted cosine compares gradients aggregated first within response groups using their advantages, then across prompts. Relative MSE divides aggregate squared error by the reference's squared norm. At each budget, we tune one FEROM configuration on the displayed cases using token-only advantage-weighted cosine, then reuse it across scoring choices and metrics. Results average 16 paired planner draws; bands show one standard deviation.

\subsection{Low-Budget Training Ablations}
\label{app:branching-protocol}
\paragraph{Training and validation.}
We train all LLaDA-8B-Instruct parameters in bfloat16 on 17,174 AceCoder examples and validate on 256 disjoint problem families. Each update uses two unique prompts with eight responses each. Rollouts use ESPO-style prompts without truncation, 256 generated tokens, block length 32, and temperature 0.9. Training reward combines supplied-test pass fractions with a format bonus. The scoring budget is 128 state evaluations per response, half of a full trajectory, and the loss is normalized by 256. All arms use conditional-CDF reveal probabilities. AdamW uses the betas and weight decay in Table~\ref{tab:training-common}, with constant learning rate \(3\times10^{-6}\) and gradient clipping at 0.2.

Validation runs before training and every ten updates using greedy confidence-based decoding at the training generation length. The curves report execution test-pass fractions without the format bonus; training curves use a trailing 50-step mean.

\paragraph{Branching ablation.}
\emph{Single trajectory (macro)} and \methodname use the same fixed-horizon macro reveals, 4,096 conditional ranking samples per evaluated state, and token-only scores. The control follows one path; \methodname permits intermediate-state branching with at most two continuations per block and at most 25\% of the block's scoring budget assigned to secondary continuations.

\paragraph{Reveal-order scoring ablation.}
We separately compare \emph{Rollout trajectory} and \emph{Rollout trajectory + order score} on actual one-token rollout transitions. Both sample states across trajectory-depth strata and apply inverse-inclusion weights. Only the second arm adds the conditional reveal-order score to the token score.

\subsection{Runtime Measurement Scope}
\label{app:timing-protocol}
\begin{figure}[htb]
    \centering
    \includegraphics[width=\linewidth]{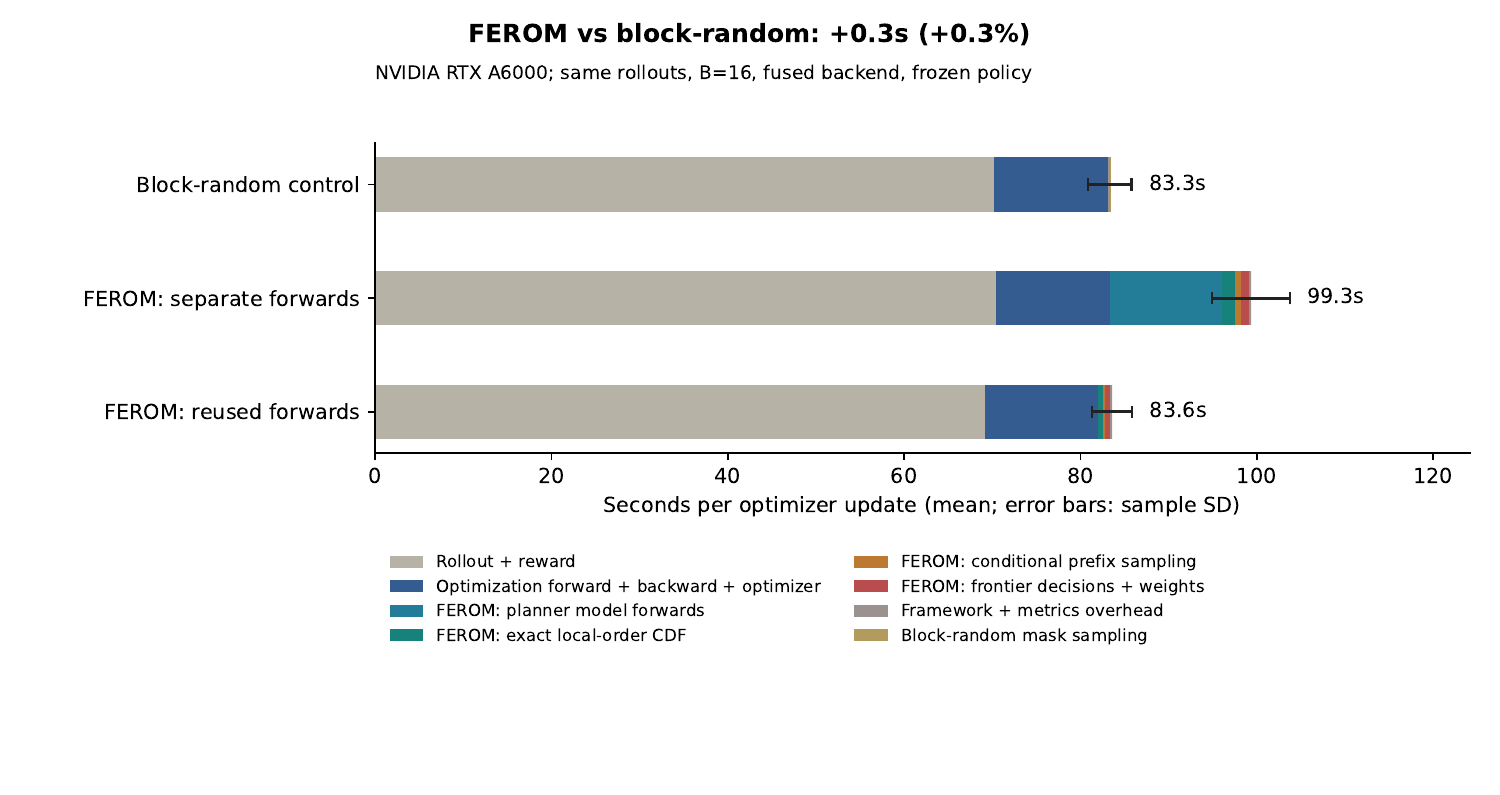}
    \caption{\textbf{Computational cost of CDF-based FEROM.} Matched block-random, memory-efficient replay (separate forwards), and shared-forward profiles. Stacked bars separate model work from CDF evaluation, macro-prefix sampling, and frontier bookkeeping. Error bars show sample standard deviations over three measured updates; the shared-forward timing difference is smaller than this variability.}
    \label{fig:time-usage}
\end{figure}

Figure~\ref{fig:time-usage} profiles LLaDA-8B-Instruct updates on one RTX A6000 using 4-bit NF4 QLoRA and bfloat16 mixed precision. Each update uses one GSM8K prompt with two 256-token responses, block length 32, temperature 0.9, and a scoring budget of 16 state evaluations per response. FEROM uses CDF-based reveal probabilities and 1,024 conditional ranking samples. All methods use fused token scoring, activation checkpointing, and zero LoRA dropout.

All methods use identical rollouts, and both FEROM implementations share their scoring states and coefficients. We hold the policy fixed with zero learning rate while measuring the complete update, including backward passes, clipping, and optimizer steps. Timings average three updates after one warm-up, with CUDA synchronization and non-overlapping phase measurements; setup, I/O, logging, and communication are excluded. Memory-efficient replay uses separate planning and scoring forwards; shared-forward execution reuses their activations.

\section{Additional Comparisons with JustGRPO}
\label{app:justgrpo-comparison}
We additionally compare with JustGRPO~\cite{ni2026flexibility}, following its full-parameter training setup in bfloat16 rather than the 4-bit QLoRA setup of the main-table experiments. We use the released JustGRPO evaluation pipeline with 256-token confidence-based generation.

\subsection{End-to-End Training from LLaDA-Instruct}
We initialize FEROM from LLaDA-Instruct and train all model parameters with 64 unique prompts per update and a half-trajectory scoring budget, \(B/T=0.5\), where \(T\) is the number of one-token updates in a full trajectory. Table~\ref{tab:justgrpo-end-to-end} compares these results with JustGRPO.

\begin{table}[htb]
    \centering
    \caption{\textbf{End-to-end training from LLaDA-Instruct.} Accuracy/pass@1 (\%) at generation length 256; JustGRPO results are from~\cite{ni2026flexibility}. Bold marks the best value in each benchmark column.}
    \label{tab:justgrpo-end-to-end}
    \small
    \begin{tabular}{lccccc}
    \toprule
    Method & \(B/T\) & GSM8K & MATH-500 & HumanEval & MBPP \\
    \midrule
    \methodname & 0.5 & 87.12 & 44.80 & 48.78 & 43.40 \\
    JustGRPO & 1.0 & \textbf{89.10} & \textbf{45.10} & \textbf{49.40} & \textbf{52.40} \\
    \bottomrule
    \end{tabular}
\end{table}

With half a trajectory's scoring budget, FEROM achieves performance close to JustGRPO on both math benchmarks and HumanEval. These runs also improve accuracy on the math benchmarks over the main-table results.

\subsection{Continual Training from JustGRPO-Code}
Starting from the released JustGRPO-Code checkpoint, we compare paired continuations with native JustGRPO and FEROM using either confidence-based or left-to-right rollouts. All continuations use full-parameter updates, 16 unique prompts with 16 responses each, and a full-trajectory scoring budget. Table~\ref{tab:justgrpo-continuation} reports the completed evaluations.

\begin{table}[htb]
    \centering
    \caption{\textbf{Continual training from JustGRPO-Code.} Pass@1 (\%) at generation length 256. Bold marks the best continuation at each update count.}
    \label{tab:justgrpo-continuation}
    \small
    \begin{tabular}{lccc}
    \toprule
    Continuation & Updates & HumanEval & MBPP \\
    \midrule
    Released checkpoint & 0 & 48.78 & 53.00 \\
    \midrule
    JustGRPO & 10 & 48.17 & \textbf{53.80} \\
    \methodname, confidence rollout & 10 & 50.00 & 53.20 \\
    \methodname, left-to-right rollout & 10 & \textbf{50.61} & 53.20 \\
    \midrule
    JustGRPO & 20 & \textbf{51.22} & \textbf{53.80} \\
    \methodname, confidence rollout & 20 & \textbf{51.22} & 52.20 \\
    \methodname, left-to-right rollout & 20 & 50.61 & 53.20 \\
    \bottomrule
    \end{tabular}
\end{table}

After ten updates, both FEROM continuations improve HumanEval and MBPP over the released checkpoint, with left-to-right rollout yielding the stronger HumanEval result. After twenty updates, confidence-rollout FEROM matches native JustGRPO on HumanEval. These results show that FEROM can further improve the HumanEval performance of a policy already trained with JustGRPO.

\paragraph{Discussion and future work.}
We hypothesize that the remaining performance gap to JustGRPO mainly reflects limited diversity of high-reward rollouts under LLaDA-style confidence-based decoding. FEROM targets the reveal-order marginal induced by a given decoder; improving the diversity and quality of its rollout proposals calls for co-design of decoding and policy optimization. This co-design is beyond the scope of the present work and will be explored in future work.

\section{Broader Impact}
\label{app:broader-impact}
FEROM is foundational research on policy optimization for masked diffusion language models, with potential benefits for reasoning and code generation. It shares the general downstream benefits and risks of language-model research; this work does not introduce a new application or deployment.

\newpage
\section*{NeurIPS Paper Checklist}

\begin{enumerate}

\item {\bf Claims}
    \item[] Question: Do the main claims made in the abstract and introduction accurately reflect the paper's contributions and scope?
    \item[] Answer: \answerYes{}
    \item[] Justification: The abstract and introduction summarize the marginal-policy formulation, finite-budget estimator, and empirical findings. Sections~\ref{sec:theory}--\ref{sec:ferom_grpo} distinguish the theoretical identity from the practical surrogate.

\item {\bf Limitations}
    \item[] Question: Does the paper discuss the limitations of the work performed by the authors?
    \item[] Answer: \answerYes{}
    \item[] Justification: Limitations are discussed in \S\ref{sec:conclusion}.

\item {\bf Theory assumptions and proofs}
    \item[] Question: For each theoretical result, does the paper provide the full set of assumptions and a complete (and correct) proof?
    \item[] Answer: \answerYes{} %
    \item[] Justification: Theorem~\ref{thm:rb_policy_gradient} states the assumptions of the marginal policy-gradient identity; Appendix~\ref{appx:theory} provides its proof and variance analysis. Appendices~\ref{app:local_order} and~\ref{app:exact_dag_dp} give the reveal-probability and occupancy derivations.

    \item {\bf Experimental result reproducibility}
    \item[] Question: Does the paper fully disclose all the information needed to reproduce the main experimental results of the paper to the extent that it affects the main claims and/or conclusions of the paper (regardless of whether the code and data are provided or not)?
    \item[] Answer: \answerYes{} %
    \item[] Justification: Sections~\ref{sec:frm}--\ref{sec:ferom_grpo} and Appendix~\ref{app:frm_variants} describe the estimator and objective. Appendix~\ref{app:experimental-protocols} specifies the datasets, training settings, evaluation protocol, and study procedures.

\item {\bf Open access to data and code}
    \item[] Question: Does the paper provide open access to the data and code, with sufficient instructions to faithfully reproduce the main experimental results, as described in supplemental material?
    \item[] Answer: \answerYes{}
    \item[] Justification: We provide the code and usage instructions at \url{https://github.com/ziansu/ferom}, also linked in the abstract. We use public datasets, with their sources and preparation described in Appendix~\ref{app:experimental-protocols} and the repository.

\item {\bf Experimental setting/details}
    \item[] Question: Does the paper specify all the training and test details (e.g., data splits, hyperparameters, how they were chosen, type of optimizer) necessary to understand the results?
    \item[] Answer: \answerYes{} %
    \item[] Justification: Appendix~\ref{app:experimental-protocols} gives data splits, training and rollout settings, optimizer hyperparameters, and evaluation procedures; Appendix~\ref{app:justgrpo-comparison} describes the additional JustGRPO comparisons.

\item {\bf Experiment statistical significance}
    \item[] Question: Does the paper report error bars suitably and correctly defined or other appropriate information about the statistical significance of the experiments?
    \item[] Answer: \answerYes{}
    \item[] Justification: Figure~\ref{fig:proxy-eval-approx} reports one standard deviation over repeated planner draws, and Figure~\ref{fig:time-usage} reports timing standard deviations over measured updates. Appendices~\ref{app:offline-protocol} and~\ref{app:timing-protocol} define these repetitions; end-to-end training results do not include independent-run error bars.

\item {\bf Experiments compute resources}
    \item[] Question: For each experiment, does the paper provide sufficient information on the computer resources (type of compute workers, memory, time of execution) needed to reproduce the experiments?
    \item[] Answer: \answerYes{} %
    \item[] Justification: Appendix~\ref{app:main-training} describes the main-table hardware and training precision. Appendix~\ref{app:timing-protocol} and Figure~\ref{fig:time-usage} give the profiling hardware, measurement settings, and computational overhead.
    
\item {\bf Code of ethics}
    \item[] Question: Does the research conducted in the paper conform, in every respect, with the NeurIPS Code of Ethics \url{https://neurips.cc/public/EthicsGuidelines}?
    \item[] Answer: \answerYes{}
    \item[] Justification: Our experiments use public models and benchmarks and do not involve human participants or new personal-data collection.

\item {\bf Broader impacts}
    \item[] Question: Does the paper discuss both potential positive societal impacts and negative societal impacts of the work performed?
    \item[] Answer: \answerYes{}
    \item[] Justification: Appendix~\ref{app:broader-impact} briefly discusses potential benefits and general downstream impacts of this algorithmic work.
    
\item {\bf Safeguards}
    \item[] Question: Does the paper describe safeguards that have been put in place for responsible release of data or models that have a high risk for misuse (e.g., pre-trained language models, image generators, or scraped datasets)?
    \item[] Answer: \answerNA{} %
    \item[] Justification: This work studies an optimization algorithm on existing public models and benchmarks, without introducing a new scraped dataset or deployment.

\item {\bf Licenses for existing assets}
    \item[] Question: Are the creators or original owners of assets (e.g., code, data, models), used in the paper, properly credited and are the license and terms of use explicitly mentioned and properly respected?
    \item[] Answer: \answerYes{} %
    \item[] Justification: We properly cite the assets we use.

\item {\bf New assets}
    \item[] Question: Are new assets introduced in the paper well documented and is the documentation provided alongside the assets?
    \item[] Answer: \answerYes{}
    \item[] Justification: We document the FEROM implementation at \url{https://github.com/ziansu/ferom}, including installation, training configurations, data preparation, and usage instructions.

\item {\bf Crowdsourcing and research with human subjects}
    \item[] Question: For crowdsourcing experiments and research with human subjects, does the paper include the full text of instructions given to participants and screenshots, if applicable, as well as details about compensation (if any)? 
    \item[] Answer: \answerNA{}
    \item[] Justification: Crowdsourcing or human subjects are not involved.

\item {\bf Institutional review board (IRB) approvals or equivalent for research with human subjects}
    \item[] Question: Does the paper describe potential risks incurred by study participants, whether such risks were disclosed to the subjects, and whether Institutional Review Board (IRB) approvals (or an equivalent approval/review based on the requirements of your country or institution) were obtained?
    \item[] Answer: \answerNA{}
    \item[] Justification: Crowdsourcing or human subjects are not involved.

\item {\bf Declaration of LLM usage}
    \item[] Question: Does the paper describe the usage of LLMs if it is an important, original, or non-standard component of the core methods in this research? Note that if the LLM is used only for writing, editing, or formatting purposes and does \emph{not} impact the core methodology, scientific rigor, or originality of the research, declaration is not required.
    \item[] Answer: \answerNA{} %
    \item[] Justification: LLM assistance is limited to routine writing, editing, and coding support, rather than an additional component of the proposed method.

\end{enumerate}

\end{document}